\pdfoutput=1
\documentclass[preprint,12pt,authoryear]{elsarticle}

\usepackage{graphicx,enumitem,amsmath,amssymb}
\usepackage{float}
\usepackage{subcaption}
\usepackage{bm}
\usepackage{algorithm,algpseudocode}
\usepackage{rotating}

\graphicspath{ {figures/} }

\usepackage{xcolor}

\newcommand{\sq}[1]{`#1'}
\newcommand{\dq}[1]{``#1''}
\DeclareMathOperator*{\argmax}{arg\,max}
\DeclareMathOperator*{\argmin}{arg\,min}

\makeatletter
\newsavebox\myboxA
\newsavebox\myboxB
\newlength\mylenA
\newcommand*\xoverline[2][0.75]{%
    \sbox{\myboxA}{$\m@th#2$}%
    \setbox\myboxB\null
    \ht\myboxB=\ht\myboxA%
    \dp\myboxB=\dp\myboxA%
    \wd\myboxB=#1\wd\myboxA
    \sbox\myboxB{$\m@th\overline{\copy\myboxB}$}
    \setlength\mylenA{\the\wd\myboxA}
    \addtolength\mylenA{-\the\wd\myboxB}%
    \ifdim\wd\myboxB<\wd\myboxA%
       \rlap{\hskip 0.5\mylenA\usebox\myboxB}{\usebox\myboxA}%
    \else
        \hskip -0.5\mylenA\rlap{\usebox\myboxA}{\hskip 0.5\mylenA\usebox\myboxB}%
    \fi}
\makeatother

\journal{Computer Speech and Language}
\begin{document}

\begin{frontmatter}

\title{An analysis of observation length requirements for machine understanding of human behaviors from spoken language}


\author[uscaddress]{Sandeep Nallan Chakravarthula\corref{mycorrespondingauthor}}
\cortext[mycorrespondingauthor]{Corresponding author}

\author[utahaddress]{Brian R.W. Baucom}
\author[uscaddress]{Shrikanth Narayanan}
\author[uscaddress]{Panayiotis Georgiou}

\address[uscaddress]{Department of Electrical and Computer Engineering, Viterbi School of Engineering, University of Southern California, Los Angeles, CA, USA}
\address[utahaddress]{Department of Psychology, College of Social \& Behavioral Science, University of Utah, Salt Lake City, UT, USA}


\begin{abstract}
The task of quantifying human behavior by observing interaction cues is an important and useful one across a range of domains in psychological research and practice.
Machine learning-based approaches typically perform this task by first estimating behavior based on cues within an observation window, such as a fixed number of words, and then aggregating the behavior over all the windows in that interaction.
The length of this window directly impacts the accuracy of estimation by controlling the amount of information being used.
The exact link between window length and accuracy, however, has not been well studied, especially in spoken language.
In this paper, we investigate this link and present an analysis framework that determines appropriate window lengths for the task of behavior estimation.
Our proposed framework utilizes a two-pronged evaluation approach: (a) extrinsic similarity between machine predictions and human expert annotations, and (b) intrinsic consistency between intra-machine and intra-human behavior relations.
We apply our analysis to real-life conversations that are annotated for a large and diverse set of behavior codes and examine the relation between the nature of a behavior and how long it should be observed.
We find that behaviors describing negative and positive affect can be accurately estimated from short to medium-length expressions whereas behaviors related to problem-solving and dysphoria require much longer observations and are difficult to quantify from language alone.
These findings are found to be generally consistent across different behavior modeling approaches.
\end{abstract}

\end{frontmatter}

\section{Introduction}
\label{sec:intro}

Human interactions involve social-cognitive abilities of varying levels of complexity such as speech detection, language understanding, emotion recognition and appropriate response generation.
Among these, the ability to reliably and accurately assess a person's behavior\footnote{we use the term \sq{behavior} to refer to not just physical actions such as facial expressions, body gestures and speech but the underlying state of mind that is expressed through these actions, and how those are perceived by domain experts} by observing their verbal and non-verbal cues is a considerably complex and important one.
Such a skill is especially important for both delivery and assessment in psychological research domains such as Couples Therapy \citep{christensen2004}, Addiction Counseling \citep{baer2009agency} and Cancer Care \citep{reblin2019communication}.
In these encounters, human experts perform formal behavioral coding by observing interactions between the provider and the client and quantitatively annotating their behavior along different dimensions, which is then used to provide feedback and improve the clinical effectiveness of care.

Subsequently, there have been efforts \citep{narayanan2013behavioral} to automate this behavior annotation (or coding) process using machine learning so that rapid and inexpensive feedback can be provided to the stakeholders.
Previous work has shown that automated coding systems are effective at quantifying behaviors as varied as \textit{Negativity} \citep{georgiou2011s,black2013toward,chakravarthula2015language,tseng2017approaching}, \textit{Depression} \citep{gupta2014multimodal,morales2018linguistically} and \textit{Empathy} \citep{xiao2012analyzing,gibson2016deep,perez2017understanding} from multimodal interaction cues.
However, there are some critical aspects of this behavior assessment process which humans can handle naturally and easily but machines still cannot.
One such aspect is the notion of how much information needs to be observed in order to reliably assess behavior, which will be the focus of investigation in this paper.

When assessing a person's behavior based on their interaction cues, humans look at factors such as the intensity of expression, context and how frequently the behavior is observed \citep{baucom2011observed}.
The latter two imply that an appropriately long window is used to observe the cues before making a judgment about the behavior; for lexical cues, we measure the length of this observation window in terms of the \textbf{number of words} spoken.

While some behaviors can be assessed based on short-duration cues, others require observations along longer time-scales.
For example, one can sense that a person is \textit{Angry} if they say something as brief as \dq{Shut up!}, but it is difficult to judge whether they are \textit{Engaged} in a discussion unless a longer and more involved conversation is observed.
Based on this, it is intuitive to expect that evaluating different behaviors would require different observation window lengths.
Such associations have been exhibited by humans when judging characteristics such as personality traits \citep{blackman1998effect}, non-verbal behaviors \citep{murphy2018predictive} and group dynamics \citep{satterstrom2019thin}.

However, it is not clear as to how these associations manifest in automated systems that quantify behaviors based on interaction cues.
Unlike emotions, which are simple and rapid \citep{baumeister2007emotion} and can be reliably estimated from short observations such as a few seconds \citep{schuller2012avec}, a sentence \citep{zadeh2018multimodal} or a speaker turn \citep{busso2008iemocap}), the rich variety of human behaviors can be much more complicated and long-ranging.
Even expert coders in the field of psychological research typically first have to be trained according to domain-specific guidelines or manuals before they can start coding patients' behaviors, such as CIRS \citep{heavey2002couples} and MITI \citep{moyers2003motivational}. 
This complexity can potentially give rise to uncertainty at the time of assessment, which then necessitates longer observations in order to achieve confident and reliable annotation.
Furthermore, the annotation time-frames for coding different behaviors can range from as short as 30 seconds \citep{heyman2004rapid} to as long as 10 minutes \citep{heavey2002couples}, demonstrating the potential variability in observation lengths.
These facets of behavior coding demonstrate the need for investigating the role of the length of observation for specific behavioral characterization.

\begin{figure}[!h]
\centering
\includegraphics[height=0.5\linewidth, width=0.99\linewidth]{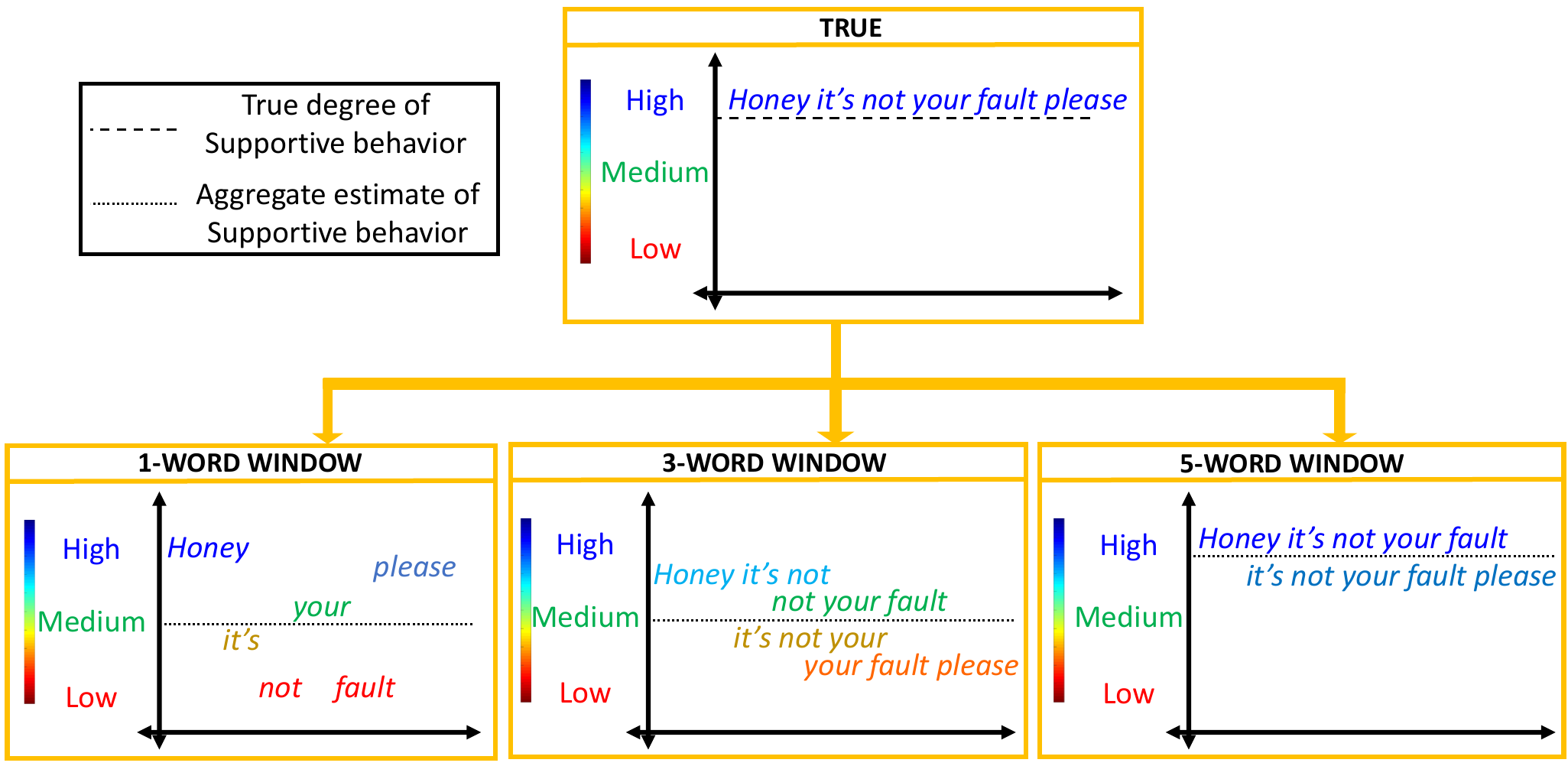}
\caption{\textit{Toy example illustrating the effect of observation window length on behavior estimation:}
The true degree of \textit{Supportive} behavior of the utterance \dq{Honey it's not your fault please} is \textbf{High}.
The system's predicted behavior is the aggregate of the behavior estimates from all the windows.
At short window lengths (1, 3 words), insufficient information leads to noisy estimates and an incorrect prediction that the degree of behavior is \textbf{Medium}.
However, as the window becomes longer (5 words), estimates are more accurate and the system correctly predicts that the degree is \textbf{High}}
\label{fig:toy_resolution}
\end{figure}

These considerations assume significant importance in applications that rely on moving-window approaches to estimate behavior.
Such approaches typically first break down an interaction into windowed segments using a fixed-length observation window, such as a fixed number of words or speaker turns.
Then, behavior estimates are computed within each window and combined over all the windows to obtain an aggregate estimate of behavior that characterizes the entire interaction.
These approaches are used in psychotherapy research where interactions, or \dq{sessions}, are often analyzed and evaluated at session-level \citep{xiao2012analyzing,gibson2016deep,tseng2017approaching}.
In such applications, the choice of length of the observation window is important; too short a window can result in noisy or incorrect estimates due to insufficient information being used, and as a result, the aggregate behavior will be inaccurate, as illustrated in the toy example in Figure~\ref{fig:toy_resolution}.
This choice also becomes important in multi-label tasks where recognizing different behaviors might necessitate the use of a different window length for each behavior.

In this work, we present a systematic analysis of the observation window length for quantifying behavior and how this varies for different behaviors.
Specifically, we are interested in empirically identifying the \textbf{minimum} amount of \textbf{language} information, measured in \textbf{number of words}, from which a behavior can be judged.
Through this analysis, we aim to address the following questions:

\begin{enumerate}[itemsep=0em]
\item Do different behaviors need observations of different lengths in order to be quantified from language?
\item How is the nature or type of behavior related to the length of its required observation window?
\end{enumerate}

Our proposed analysis framework consists of two components: (1) A pair of evaluation metrics that describe the window-level and interaction-level quality of behavior predictions by the system at a given window length, and (2) A step-by-step procedure that progressively examines how these metrics for a given behavior change as the window length is varied, based on which the appropriate window length is determined.
We conduct our analysis on the Couples Therapy corpus \citep{christensen2004} which contains real-life interactions coded for a large and diverse set of behaviors in human interaction.
We then compare our findings against existing literature on similar behavior analyses using modalities such as audio, as well as work in psychology related to constructs such as affect and personality traits.

This paper is structured as follows:
In Sec.~\ref{sec:related}, we describe existing work in psychology and machine learning that has dealt with related problems, following which we formally define our problem of interest in Sec.~\ref{sec:problem}.
We then introduce our proposed analysis metrics and procedure in Sec.~\ref{sec:analysis} and explain how they address the problem of determining appropriate observation window lengths for behaviors.
Sec.~\ref{sec:data} then describes the Couples Therapy dataset, followed by details of our experimental setup in Sec.~\ref{sec:experiment}.
Finally, in Sec.~\ref{sec:results}, we present our findings and discuss their implications for the research questions posed earlier.
We conclude in Sec.~\ref{sec:conc_futwor} and comment on potential improvements and extensions to this work.

\section{Related Work}
\label{sec:related}

A body of work in psychology that is related to, but not the same as, our notion of \dq{window length} is the one which studies \textit{Thin Slices} of observed behavior \citep{ambady1992thin}.
It refers to excerpts or snippets of an interaction that can be used to arrive at a similar judgment of behavior to as if the entire interaction had been used.
Essentially, it implies that an entire interaction can be replaced with just a windowed part (which is different from our aim of identifying the best window through which to view the \textbf{entire interaction}).
The effect of the location of these slices has been investigated as well; the conventional approach is to situate the slices near the start of the interaction.
The effectiveness of thin slices has been observed in many applications, ranging from judging personality traits \citep{blackman1998effect,krzyzaniakeffect} such as the \dq{big five} \citep{mccrae1986evaluating} to viewer impressions of TED talks \citep{cullen2017thin} such as \dq{funny} and \dq{inspiring}.

Notably, Carney et al. \citep{carney2007thin} studied the accuracy of impressions for Affect, Extraversion and Intelligence at different thin-slice durations, locations, etc.
Accuracy was measured as the correlation between the true value of a construct (whether rated or self-reported) and the impression based on the thin slice, and it was observed that, in general, accuracy increased as the slice length increased from 5 seconds to 5 minutes.
Furthermore, they found that \textit{Negative} affect could be assessed with similar accuracies at all slice lengths whereas \textit{Positive} affect was best assessed only when thicker slices were used.
These works provide an encouraging support for analyzing the window length of behaviors along similar lines.

There has also been a great deal of work in psychology on studying how humans perceive and process events characterized by concepts such as \textit{good} and \textit{bad}.
Specifically, there exists a notion that \dq{bad} is \dq{stronger} than \dq{good} \citep{baumeister2001bad}, meaning that undesirable or unpleasant events have a greater impact on an individual than desirable, pleasant ones.
The behavior constructs that we are interested in analyzing are similar to concepts that have been shown to exhibit this phenomenon in previous works.
For instance, Oehman et al. \citep{ohman2001face} found that people detected threatening faces more quickly and accurately than those depicting friendly expressions.
In a similar experiment, Krull et al. \citep{krull1998smiles} showed videos of either happy or sad individuals to participants and reported that happiness evoked more spontaneous inferences while sadness drew slower ones.
This shows that different concepts are perceived differently, depending on their valence; hence, in this work, we investigate how the nature of different behavior constructs is tied to aspects of their expression and perception in language such as window length, aggregation mechanism, etc.

Some approaches in machine learning and speech processing that are similar to ours have investigated the accuracy of behavior prediction using acoustic vocal cues.
Xia et al. \citep{xia2015dynamic} found that as the observation window used to compute acoustic features was increased from 2 seconds to 50 seconds, the classification accuracy generally improved, with \textit{Negative} and \textit{Positive} behaviors gaining the most.
Similar results were reported by Li et al. \citep{li2020linking} who classified behaviors such as \textit{Acceptance}, \textit{Negativity} and \textit{Sadness} by employing emotion-based behavior models on acoustic features.
In their models, the receptive field (measured in seconds) of a 1-D Convolutional Neural Network-based system served as the observation window for vocal cues.
In general, they found that behaviors relating to negative affect, such as \textit{Negativity}, were classified more accurately than behaviors such as \textit{Acceptance} and \textit{Sadness}.
They also observed that increasing the receptive field from 4 seconds to 64 seconds generally resulted in better classification, with \textit{Sadness} performing best at 16 seconds while \textit{Negativity} performed best at 64 seconds.
Other efforts have addressed related aspects; for instance, Lee et al. \citep{lee2012based} examined whether the behavior annotation process is driven more by a gradual, causal mechanism or by isolated salient events, which imply the use of long and short observation windows respectively.

While these works have contributed to a better understanding of the effect of observation windows, they are limited in the variety of constructs that are analyzed.
Furthermore, they mostly focus on acoustic and vocal cues and not enough on the lexical characteristics.
Hence, the novelty of our work lies in (1) analyzing the effect of observation lengths in the lexical modality and (2) performing this analysis using a large and diverse set of real-life human behavior constructs.
Through our analysis, we aim to understand the relation between the nature of the behavior and how long an automated system needs to observe its expression in language in order to accurately estimate it.

\section{Problem Statement}
\label{sec:problem}

\begin{figure}[!h]
\centering
\includegraphics[height=0.23\linewidth, width=0.99\linewidth]{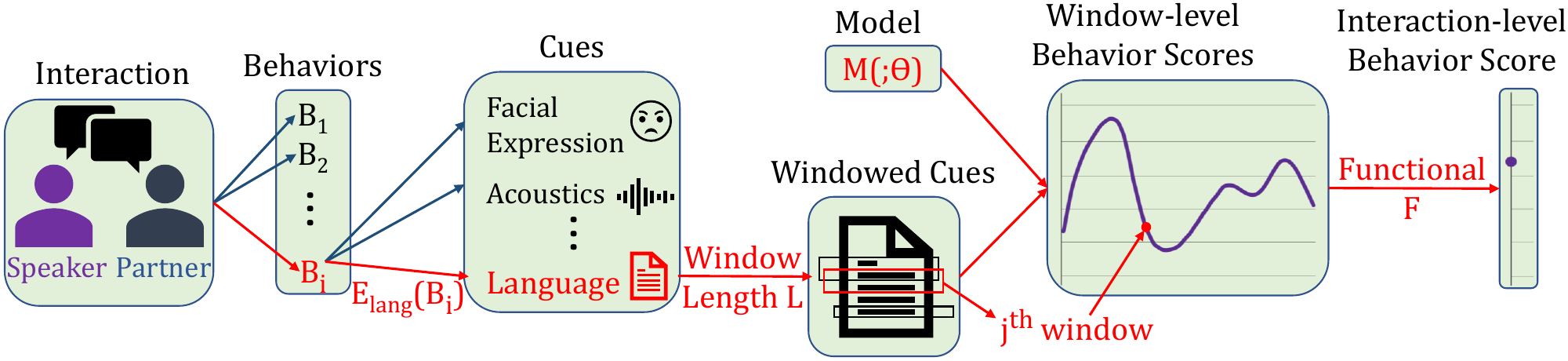}
\caption{\textit{Automated quantification of behavior from lexical cues using a moving-window approach:}
During an interaction, interlocutors express behaviors such as $B_i$ through conversational cues such as language cues.
The text of the conversation is decomposed into windowed chunks, each chunk \textit{L} words long.
Then, model \textit{M} is used to score the text inside each window, resulting in a trajectory of window-level scores.
Finally, a functional \textit{F} is applied on the trajectory to obtain a summary of the behavior during the interaction}
\label{fig:beh_quant}
\end{figure}

Figure~\ref{fig:beh_quant} depicts a typical system setup used in previous works \citep{xiao2012analyzing,gibson2016deep,tseng2017approaching} that employed a moving-window approach for estimating behaviors.
Since our proposed analysis assumes such a setup, we will use its components to formally describe the problem of determining the appropriate observation window length for estimating behaviors.

Let's suppose that we want to estimate behavior $B_i$ using a set of observed data samples \textit{D}, where each sample is a sequence of words.
Let $A_i \in \mathbb{R}^{|\textit{D}|}$ be the ground-truth annotations of $B_i$ in \textit{D} and let \textit{C} be a metric that is used to evaluate the estimation results against $A_i$; the higher the value of \textit{C}, the more accurate our estimates are.

Let $E_{lang}(B_i)$ represent the degree to which the behavior $B_i$ is actually expressed in language.
This is not known beforehand, so we assume it to be high and that it is possible to estimate $B_i$ from language.
Let \textit{M} denote a machine learning model (e.g., Deep Neural Network) that estimates a scalar value from a sequence of words and $\theta$ represent its learnable parameters (e.g., weights).
Finally, let $L$ represent the window length at which $B_i$ is observed in language and $F$ denote a statistical functional that maps a sequence of scalar values to a single scalar value. Then, the quality of behavior quantification can be expressed as:
\begin{equation}
\mathbb{Q}_i = C(F(M(E_{lang}(B_i), L, D; \theta)), ~A_i) \label{eqn:q}
\end{equation}

\noindent Our goal is to identify the window length $L_i^*$ that maximizes $\mathbb{Q}_i$ for $B_i$:
\begin{align}
L_i^* = \argmax\limits_{L}\text{ }\mathbb{Q}_i \label{eqn:q_max}
\end{align}

We argue that $\mathbb{Q}_i$ can be high only when $E_{lang}(B_i), L, M, \theta$ and $F$ are all appropriate together. If even one of them is flawed or incompatible with the rest, then it would adversely affect $\mathbb{Q}_i$, as explained below:
\begin{itemize}
\item $E_{lang}(B_i)$: The behavior $B_i$ must be sufficiently expressed in the lexical channel to begin with; otherwise, it might not be possible to observe it using lexical cues alone (for example, if it is instead primarily expressed through nonverbal vocal cues such as laughter).
In general, we do not have information about $E_{lang}(B_i)$ beforehand; instead, we simply assume that it is high enough so that it is possible to estimate $B_i$ from language.
\item $L$: The observation window must be long enough to observe $B_i$; otherwise, the incomplete information from partial observations can lead to noisy or incorrect estimates.
\item $M$: The model must be well-suited for capturing $B_i$.
For example, quantifying a behavior that is based on the actions of both speakers requires a model that looks at both speakers; using a single-speaker model instead would result in inaccurate estimates due to insufficient information.
\item $\theta$: The model must be well-trained; otherwise, its estimates might be inaccurate. This is dependent on the training process, the amount and quality of data used to estimate parameters, etc.
\item $F$: The aggregating functional must be well-suited for summarizing $B_i$; otherwise, the resulting aggregate estimate might not match the ground-truth annotations $A_i$.
For example, the functional \textit{mode}, which identifies the most frequently occurring value, might not be appropriate for summarizing a behavior that is expressed very infrequently.
\end{itemize}

We reason that a high value of $\mathbb{Q}$ is indicative of all the aforementioned factors being appropriate and a low value is indicative of a limitation in at least one of these factors.
Based on this, we now proceed to analyze the variation in $\mathbb{Q}$ for different behaviors as window length $L$ is varied.

\section{Proposed Analysis Methodology}
\label{sec:analysis}

In this section, we describe in detail our proposed framework of behavior window length analysis.
We first describe two evaluation metrics that quantify how well the system is able to accurately estimate a behavior.
Following this, we present a step-by-step procedure that progressively employs both metrics to determine the appropriate window lengths for each behavior.

\subsection{Metrics}
\label{ssec:metrics}

\subsubsection{Behavior Construct Similarity}
\label{sssec:analysis_bcs}

\begin{figure}[!h]
\centering
\includegraphics[height=0.35\linewidth, width=0.99\linewidth]{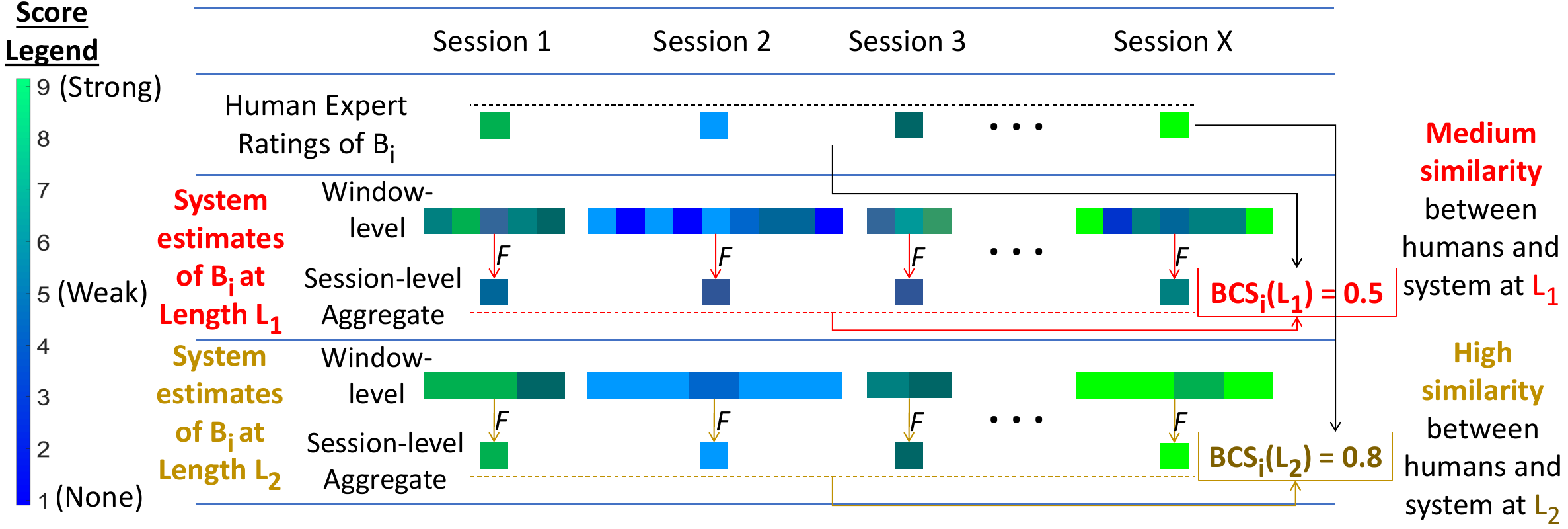}
\caption{\textit{Computation of Behavior Construct Similarity (BCS) for behavior $B_i$ at two observation window lengths $L_1$ and $L_2$:}
The session-level aggregates are more highly correlated with human expert ratings at $L_2$ (0.8) than at $L_1$ (0.5).
Therefore, $L_2$ is considered to be more appropriate than $L_1$ for estimating $B_i$}
\label{fig:bcs_example}
\end{figure}



Our first proposed metric, the Behavior Construct Similarity (BCS), measures how well the system matches humans in terms of understanding the overall, aggregate behavior content of the entire interaction.
The higher the value, the more similar the system is to humans and, hence, the more reliable the behavior estimates are.

For behavior $B_i$ and window length $L_j$, it is computed as:
\begin{equation}
BCS_i(j) = R(F(M(E_{lang}(B_i), L_j, D; \theta)), ~A_i) \label{eqn:bcs}
\end{equation}
where $R$ refers to the Spearman Correlation between human annotations $A_i$ and the system estimates, which are expected to be ordinal variables, in general.
For the functional \textit{F}, we test 3 statistics: \textit{median}, \textit{minimum} and \textit{maximum};  the former provides a useful, \dq{average} summary of behavior, as shown in previous works \citep{tseng2016couples,tseng2018honey} while the latter two represent outlier events such as large/small.

As can be seen from Eqn.~\ref{eqn:bcs}, BCS is a direct implementation of Eqn.~\ref{eqn:q}, with Spearman's Correlation \textit{R} as the choice of the evaluation metric \textit{C}.
It is computed at the session-level using Algorithm~\ref{alg:bcs} in \ref{sec:append_alg} and is, thus, a session-level measure of the system performance.
It takes values in the range [-1, 1], where -1 represents \textit{no similarity} while 1 represents \textit{full similarity}.
Figure~\ref{fig:bcs_example} shows an example of how BCS can be used to determine the appropriate window length for behavior $B_i$.


While BCS provides the best validation possible for system estimates by directly comparing them against human judgments, it nevertheless suffers from two limitations:
\begin{enumerate}
\item A poor choice of the functional $F$ in Eqn.~\ref{eqn:bcs} can result in inaccurate aggregates and, hence, low BCS, even if a sufficiently long window is used.
In such a scenario, relying on BCS alone would lead to an incorrect conclusion that the window length is not appropriate.
\item BCS cannot be quickly computed at any arbitrary window length, since we first require human annotations at that window length to compare against.
This requirement becomes practically infeasible when analyzing a large set of window lengths.
\end{enumerate}
In order to account for these, we propose an additional metric that can be quickly computed at any arbitrary window length and which does not rely on any aggregation, as described below.


\subsubsection{Behavior Relationship Consistency}
\label{sssec:analysis_brc}

Our second proposed evaluation metric, the Behavior Relationship Consistency (BRC), measures how well the system matches humans in terms of perceptual relations between different behaviors.
These refer to notions of similarity and dissimilarity that arise between different constructs due to the way they are defined; for example, \textit{Happiness} is considered similar to \textit{Joy} and \textit{Satisfaction} but opposite to \textit{Sadness}.

Thornton et al. \citep{thornton2017mental} studied emotional transitions and found that similar emotions (e.g. \textit{Anger} and \textit{Joy}) frequently tend to co-occur whereas dissimilar ones (e.g. \textit{Anger} and \textit{Disgust}) do not.
We expect behaviors to exhibit such phenomena as well and reason that a system that can accurately model related behaviors would produce estimates that also consistently reflect these relations.
The BRC metric measures this consistency; the higher its value, the greater the consistency and, thus, the more reliable the behavior estimates are.

\begin{figure}[!h]
\centering
\includegraphics[height=0.325\linewidth, width=0.99\linewidth]{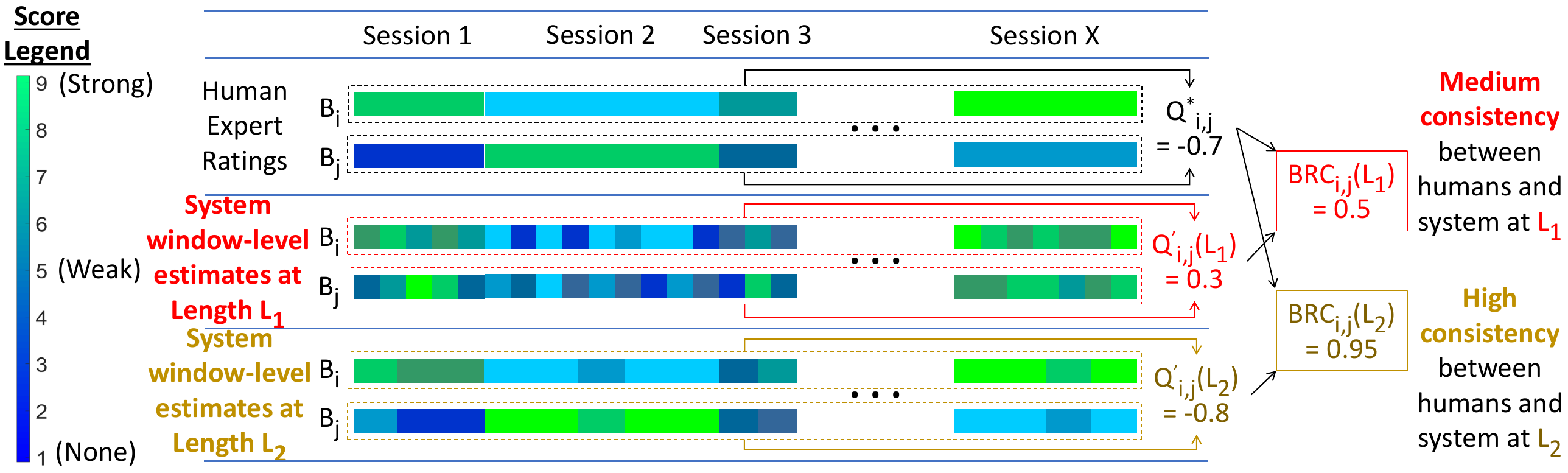}
\caption{\textit{Computation of Behavior Relationship Consistency (BRC) between behaviors $B_i$ and $B_j$ at two observation window lengths $L_1$ and $L_2$:}
The correlation between $B_i$ and $B_j$'s estimates are more similar to the correlation between $B_i$ and $B_j$'s human ratings (-0.7) at $L_2$ (-0.8) than at $L_1$ (0.3).
Therefore, $L_2$ is considered to be more appropriate than $L_1$ for estimating $B_i$ and $B_j$}
\label{fig:brc_example}
\end{figure}

BRC is defined for a pair of behaviors \{$B_i$, $B_j$\} and measures how close the Spearman Correlation between their window-level estimates at window length $L_k$ is to the Spearman Correlation between their ground-truth annotations.
It is calculated as:
\begin{align}
BRC_{i,j}(k) & = 1 - \frac{|Q_{i,j}^{*} - Q_{i,j}^{'}(k)|}{2}\label{eqn:brc}  \\
\text{where } Q_{i,j}^{*} & = R(A_i, A_j)\nonumber\\
\text{and } Q_{i,j}^{'}(k) & = R(M(E_{lang}(B_i), L_k, D; \theta), ~M(E_{lang}(B_j), L_k, D; \theta))\nonumber
\end{align}
BRC is computed at the window-level using Algorithm~\ref{alg:brc} in \ref{sec:append_alg} and is, thus, a window-level measure of the system performance.
It takes values in the range [0, 1], where 0 represents \textit{no consistency} while 1 represents \textit{full consistency}.
Figure~\ref{fig:brc_example} shows an example of how BRC can be used to determine the appropriate window length for behaviors $B_i$ and $B_j$.

It can be seen that $Q_{i,j}^{'}$ in Eqn.~\ref{eqn:brc} is similar to $\mathbb{Q}_i$ from Eqn.~\ref{eqn:q}; while $\mathbb{Q}_i$ evaluates against human annotations, $Q_{i,j}^{'}$ evaluates against estimates of other behaviors.
Thus, the effectiveness of $B_i$'s BRC is directly dependent on $B_j$'s estimates; the more accurate they are, the more we can rely on $B_i$'s BRC with $B_j$.
Using this principle, given multiple behavior pairs \{$B_i, B_j\ \forall j \in J$\}, $B_i$'s \textit{weighted} BRC is calculated as a weighted sum of its individual BRCs with $B_j$, proportional to their BCS:
\begin{align}
BRC_{i}(k) & = \sum_{j \in J} \alpha_j(k) BRC_{i,j}(k) \label{eqn:brc_multi}  \\
\text{where } \alpha_j(k) & = \frac{BCS_j(k)}{\sum\limits_{m \in J} BCS_m(k)} \label{eqn:rel_weights}
\end{align}

\begin{figure}[h!]
\centering
\includegraphics[height=1.1\linewidth, width=0.99\linewidth]{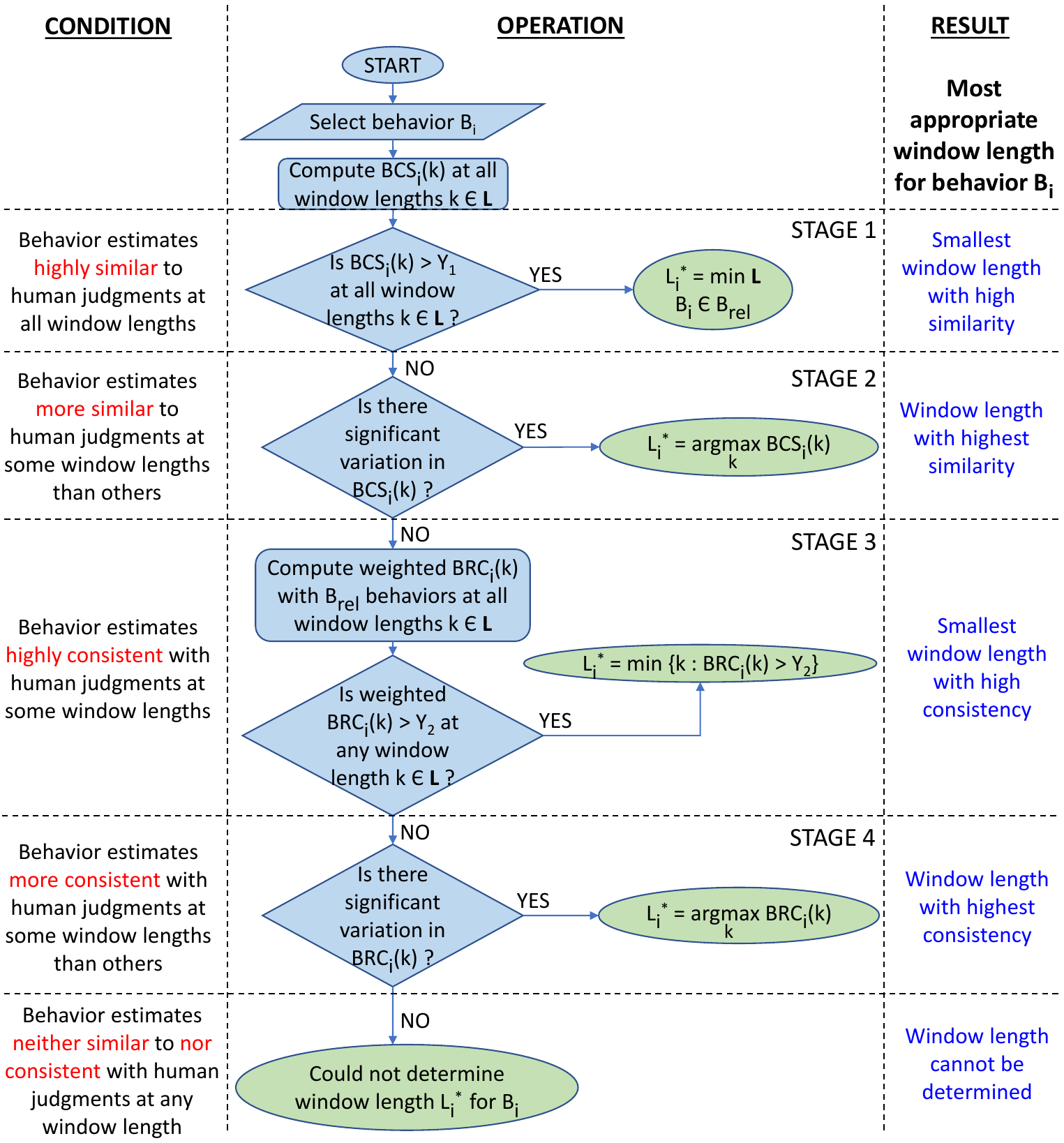}
\caption{\textit{Flowchart of the analysis procedure for determining appropriate window length of target behavior $B_i$:}
In each stage, we check if the system's estimates of $B_i$ satisfy a condition, summarized under \dq{CONDITION} and denoted in blue boxes under \dq{OPERATION}.
If satisfied, we determine the appropriate window length as denoted in green boxes under \dq{OPERATION} and summarized under \dq{RESULT}.
If the condition is not satisfied, we simply proceed to the next stage.
This procedure continues for 4 stages, beyond which we cannot make determinations about $B_i$'s window length.}
\label{fig:analysis_flow}
\end{figure}

\subsection{Procedure}
\label{ssec:proc}

With the analysis metrics defined, we now proceed to employ them in the following multi-stage fashion, depicted in the flowchart in Figure~\ref{fig:analysis_flow}.
\begin{itemize}

\item \textbf{Stage 1}:
First, we focus on those behaviors which the system can estimate with high accuracy.
As argued in Sec.~\ref{sec:problem}, if a behavior $B_i$ can be accurately estimated at window length $L$, then this implies that $L$ is appropriate for $B_i$.
As mentioned in Sec.~\ref{sec:intro}, the appropriate window length $L_i^*$ we are interested in is the minimum amount of words required to capture $B_i$.
Therefore, we identify all window lengths \textbf{L} at which its BCS, defined in Eqn.~\ref{eqn:bcs}, is high and simply select the shortest one.
\begin{align*}
\text{Identify } B_i &: BCS_i(k) > Y_1 ~ \forall k \in \textbf{L}\\
\text{Window length } L_i^* &= min \{ ~ \textbf{L} ~ \}
\end{align*}
$Y_1$ is a pre-defined threshold; the closer this threshold is to 1, the more confident we can be that the estimate of a behavior is indeed similar.
We refer to the set of behaviors identified in Stage 1 as \dq{reliable} behaviors, denoted using $B_{rel}$.

\item \textbf{Stage 2}:
Next, we examine those behaviors whose estimates are not highly similar to human judgment (i.e. BCS lower than $Y_1$) but which nevertheless are more similar at some window lengths than others.
While low similarity can be attributed to factors other than window length, since we do not vary them, any change must be solely due to the window length.
Therefore, for such behaviors, we check if their BCS shows a significant fluctuation over the entire range of window lengths, and simply pick the length at which it was highest.
\begin{gather*}
\text{Identify } B_i : max\{BCS_i(k)\} > min\{BCS_i(k)\} \text{ (significant, } p < 0.05) \\
\text{Window length } L_i^* = \argmax_k BCS_i(k)
\end{gather*}
We check the change in BCS for statistical significance by calculating the 95\% confidence interval for differences in dependent overlapping correlations using Zou's method \citep{zou2007toward}, as recommended by Diedenhofen et al. \citep{diedenhofen2015cocor}.
The change in BCS is considered to be statistically significant if the interval does not contain 0; else, it is not significant.

\item \textbf{Stage 3}:
In Stage 3, we analyze those behaviors whose estimates were not found to be similar to human judgments, as evinced from their low values of BCS.
As explained in Sec.~\ref{sssec:analysis_bcs}, however, a low value of BCS for a behavior $B_i$ does not automatically imply that its window-level estimates themselves are inaccurate; rather, it might be due to an inappropriate aggregating functional used in computing BCS.

Therefore, we inspect the window-level estimates directly by examining how consistent they are, given by the weighted BRC, defined in Eqn.~\ref{eqn:brc_multi}.
We compute the weighted BRC of $B_i$ with respect to the $B_{rel}$ behaviors that were identified in Stage 1 as being accurately estimated. 
We then check if it is higher than a pre-defined threshold $Y_2$ and pick the shortest window length at which this is true; as before, the closer $Y_2$ is to 1, the more confident we are that the estimate of a behavior is consistent.
\begin{gather*}
\text{Identify } B_i : \exists ~ k \in \textbf{L} : BRC_i(k) > Y_2 \\
\text{Window length } L_i^* = min\{ ~ k \in \textbf{L} : BRC_i(k) > Y_2 \}
\end{gather*}

\item \textbf{Stage 4}:
Lastly, we inspect behaviors whose estimates are not highly consistent (i.e. BRC lower than $Y_2$) but which nevertheless show significantly more consistency at some window lengths than others.
Therefore, similar to Stage 2, we check if their BRC varies significantly over the entire range of window lengths, and simply pick the length at which it was highest.
\begin{gather*}
\text{Identify } B_i : max\{BRC_i(k)\} > min\{BRC_i(k)\} \text{ (significant, } p < 0.05) \\
\text{Window length } L_i^* = \argmax_k BRC_i(k)
\end{gather*}
We check the change in BRC for statistical significance by calculating the 95\% confidence interval for differences in independent correlations using Zou's method \citep{zou2007toward}, as recommended by Diedenhofen et al. \citep{diedenhofen2015cocor}.

\item \textbf{End}:
For behaviors that show neither similarity nor consistency in their estimates with human judgments, we do not analyze them any further and simply conclude that we are unable to make any determinations at this point about their appropriate window lengths.

\end{itemize}

\section{Behavior Dataset}
\label{sec:data}

We now proceed to describe the Couples Therapy dataset on which we apply our analysis.
In general, the analysis framework described above can be applied in any domain where the task of interest is to analyze human-centered interactions for perceptual constructs including, but not limited to, emotion and sentiment.

The Couples Therapy project \citep{christensen2004} involved 134 real-life chronically distressed couples that attended marital therapy over a period of up to 1 year.
Its dataset consists of hundreds of real-life interactions as well as a rich and diverse set of human-annotated codes characterizing the behavior of the participants in these interactions.

\begin{center}
\begin{table*}[ht]
\centering
\setlength\doublerulesep{1em} 
\scalebox{0.62}{
\begin{tabular}{| p{4cm} | p{17cm} |}
\hline
\textbf{CIRS2 Code} & \textbf{Description} \\ \hline
Acceptance & Indicates understanding, acceptance, respect for partner’s views, feelings and behaviors \\ \hline
Perspective & Tries to understand partner’s views, feelings by clarifying and asking to hear them out \\ \hline
Responsibility & Implies self-power over feelings, thoughts, behaviors on issue being discussed \\ \hline
External & Softens criticism of partner by attributing their undesirable behaviors to external origins \\ \hline
Define & Articulates problems clearly, facilitates everyone's participation in problem solving process \\ \hline
Solution & Suggests specific solutions that could solve the problem \\ \hline
Negotiates & Offers compromises or bargains \\ \hline
Agreement & States terms of agreement, willingness to follow them with partner \\ \hline
Blame & Blames, accuses, criticizes partner and uses critical sarcasm and character assassinations \\ \hline
Change & Requests, demands, nags, pressures for change in partner \\ \hline
Withdrawal & Generally non-verbal, becomes silent, refuses to respond, discuss, argue, defend \\ \hline
Avoidance & Minimizes importance and denies existence of problem, diverts attention, delays discussion \\ \hline
Discussion & Discusses problem, shows engagement, interest and willingness in discussing issue \\ \hline \hline
\textbf{SSIRS Code} & \textbf{Description} \\ \hline
Positive & Overtly expresses warmth, support, acceptance, affection, positive negotiation \\ \hline
Negative & Overtly expresses rejection, defensiveness, blaming, and anger \\ \hline
Anger & Expresses anger, frustration, hostility, or resentment during the interaction \\ \hline
Belligerence & Quarrels, argues, verbalizes nasty comments and mean rhetorical questions \\ \hline
Disgust & Shows disregard, scorn, lack of respect and makes patronizing and insulting comments \\ \hline
Sadness & Cries, sighs, speaks in a soft or low tone, expresses unhappiness and disappointment \\ \hline
Anxiety & Expresses discomfort and stress, answers with short yes/no responses without elaboration \\ \hline
Defensiveness & Deflects criticism by defending self, accusing partner of similar behavior \\ \hline
Affection & Expresses warmth and caring for partner, speaks warmly, uses endearments \\ \hline
Satisfaction & Feels satisfaction about how topic of discussion is defined, discussed, and resolved \\ \hline
Dominance & Commands course of interaction, dominates conversation, changes subject frequently \\ \hline
Solicits Suggestions & Shows interest in and seeks partner’s suggestions, help in handling issue \\ \hline
Instrumental Support & Offers positive advice for clear, concrete actions to support partner \\ \hline
Emotional Support & Emphasizes feelings, builds confidence, and raises self-esteem in partner \\ \hline
\end{tabular}}
\caption{\textit{Description of behavior codes in Couples Therapy corpus}}
\label{tab:beh_list}
\end{table*}
\end{center}
\vspace{-1cm}

The corpus consists of audio-visual recordings, with manual transcriptions, of husband-wife couples discussing topics of marital distress in 10-minute interactions.
Each couple had at least 2 interactions or \dq{sessions}, once with each participant leading the discussion on a topic of their choice, and the total number of sessions per couple ranged from 2 to 6.
In each session, both the husband and the wife were rated for a total of 13 CIRS \citep{heavey2002couples} and 18 SSIRS \citep{jones1998couples} behavior codes by trained human annotators with a sense of what \dq{typical} behavior is like during these interactions.
The annotators were asked to observe both verbal and nonverbal expressions when rating each behavior independently and in many cases, different annotators rated different behaviors.
Each behavior in each session was rated by 3 to 9 annotators, with most of them being rated by 3 to 4.
The rating was done on a Likert scale from 1 to 9, where 1 represents \dq{absence of behavior}, 5 represents \dq{some presence of behavior} and 9 represents \dq{strong presence of behavior}.
More details about the recruitment, data collection and the annotations can be found in \citep{christensen2004,baucom2011observed}.

Consistent with previous work \citep{lee2010quantification, georgiou2011s, black2013toward, lee2014computing, tseng2017approaching}, for each speaker and behavior, we take the average of the annotators' ratings as the true rating in that session.
Therefore, for each speaker in every session, we have the manual transcription of their utterances and their behavior ratings in that session.
We disregard 4 irrelevant codes such as \dq{Is the topic of discussion a personal issue ?} and \dq{Is the discussion about husband's behavior ?}, since they are tied more to the topic of interaction and less to the speaker's behavior.
The resulting set of 27 behaviors that will be analyzed in this work are listed in Table~\ref{tab:beh_list} and categorized as follows:
\begin{itemize}[itemsep=0em]
\item \textit{Couples Interaction Rating System 2} (CIRS2): This set contains 13 codes that describe a speaker when interacting with their partner about a problem.
\item \textit{Social Support Interaction Rating System} (SSIRS): This set contains 14 codes that measure emotional features and ratings of the interaction.
\end{itemize}
We only consider those sessions where both speakers were rated for all 27 behaviors, resulting in 1325 sessions in total.
Since the content and nature of interaction vary from one couple to another, the number of words spoken by a speaker during a session ranges from around 50 to 2500, with a mean of 805 words and a standard deviation of 305.
%

\section{Experimental Setup}
\label{sec:experiment}

This section provides the details of our experimental setup, starting with the process for creating windowed samples of language from the session transcripts.
Then, we describe the machine learning model used to score conversational language for different behaviors and provide its implementation details.
Finally, we describe a perceptual grouping scheme for the behaviors being analyzed in our work to assist in the interpretability of our results.

\subsection{Windowed Scoring of Text}
\label{ssec:windows}

Let's suppose that the text transcript $T_p$, from the $p^{th}$ session, contains $O_p$ words.
Using an observation window of length $L_k$, we first decompose it into its constituent windows.
If $O_p > L_k$, then we get $O_p - L_k + 1$ windows, each one containing $L_k$ words; else, we get just one window.
Then, using model $M$, we estimate the behavior within each window independently .
Assuming that $T_p={w_1,w_2,\ldots,w_{O_p}}$, there are $O_p$ possible lengths at which it can be windowed, as shown in the first column of Table~\ref{table:resolution}.

\begin{table}[!h]
\centering
\footnotesize
\begin{tabular}{|c|c|c|c|}
\hline
\textbf{Observation} & \textbf{Window} & \textbf{No. of} & \textbf{Resolution}
\\
\textbf{Window} & \textbf{Decomposition} & \textbf{Windows} & \\ \hline
session-length & \{$w_1$,$w_2$, ... $w_{O_p}$\} & 1 & Very Coarse \\
$O_p$-1 word & \{$w_1$,$w_2$, ... $w_{{O_p}-1}$\}, \{$w_2$,$w_3$, ... $w_{O_p}$\} & 2 & Coarse \\
\multicolumn{4}{|c|}{...} \\
2-word & \{$w_1$,$w_2$\}, \{$w_2$,$w_3$\}, ... \{$w_{{O_p}-1}$,$w_{O_p}$\} & $O_p$-1 & Fine \\
1-word & \{$w_1$\}, \{$w_2$\}, ... \{$w_{O_p}$\} & $O_p$ & Very Fine \\ \hline
\end{tabular}
\caption{\textit{All possible window lengths at which an utterance with $O_p$ words can be scored}}
\label{table:resolution}
\end{table}

The window with the coarsest resolution is the \dq{session-length} window since it views the entire session as a whole, resulting in a single behavior estimate, or \dq{score}, for $T_p$.
On the other hand, the \dq{1-word} window provides the finest resolution possible since a score is generated for each word in the session, resulting in a trajectory of scores for $T_p$.
In this work, we test the following observation window lengths: \{\textit{3, 10, 30, 50, 100}\} where \textit{N} represents a window that is \textit{N} words long.
\textit{3} and \textit{10} can be qualitatively thought of as \dq{short} windows, \textit{30} and \textit{50} as \dq{medium}-length windows and \textit{100} as \dq{long} window.

The number of scores resulting from each session depends on the window length used; the longer the window, the fewer the number of scores.
For instance, since there are 1325 sessions in our dataset, estimating behavior with a \dq{session-length} window would result in one score per session and, thus, 1325 scores in total.
On the other hand, estimating with a 1-word window results in 1067727 scores in total, over all the sessions.
In our work, the window length varies from \textit{3} to \textit{100} and, hence, the total number of scores varies from 1065077 to 936738 respectively.

\subsection{Behavior Model}
\label{ssec:beh_model}

\subsubsection{Model Description}
\label{sssec:mle_model}

\begin{figure}[!h]
\centering
\includegraphics[height=0.5\linewidth, width=0.9\linewidth]{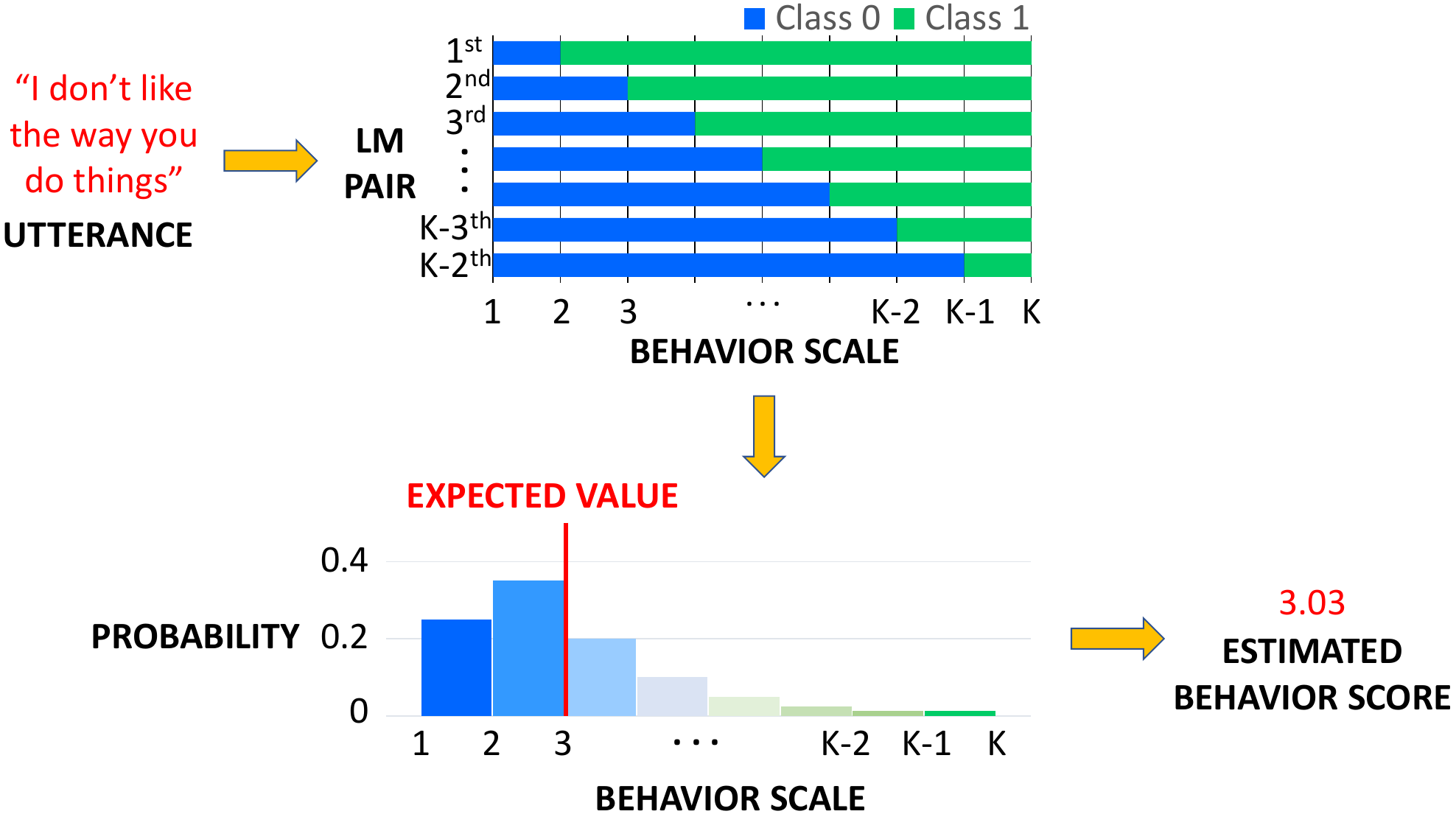}
\caption{\textit{N-gram model used to estimate behavior of a sample utterance:}
Pre-built sets of K-2 binary LMs provide likelihoods for the utterance on a behavior scale from 1 to K (in our data, K=9).
Posterior probabilities are then calculated and the expected value of the resulting distribution is used as the estimated behavior score of the utterance}
\label{fig:ngram_model}
\end{figure}

We use a Maximum Likelihood method closely following \citep{frank2001simple,rozgic2011estimation} which employs N-gram Language Models (LMs) in a cumulative fashion as shown in Figure~\ref{fig:ngram_model}.
An N-gram LM takes as input a text sequence - for example, an utterance $W$ consisting of $O$ words $W$$=$$\{w_1,w_2,\ldots,w_O\}$ - and outputs a likelihood probability given by:
\begin{align}
\label{eq:joint_prob}
P(W) = P(w_1,w_2,\ldots,w_O) \approx \prod_{n=1}^{O}P(w_n|w_{n-1},w_{n-2},...w_{n-N+1})
\end{align}
We use N-gram LMs since they have been shown to be accurate at behavior estimation in previous works \citep{georgiou2011s,chakravarthula2015assessing} and are easy and simple to train.

First, we create multiple binary partitions of our dataset based on its behavior ratings, which range from 1 to 9.
The $r^{th}$ partition consists of Class \textit{0}, which contains sessions with ratings in the range [\textit{1,r+1}], and Class \textit{1}, which contains those in the range (\textit{r+1,9}].
Then, we build a binary classifier LM pair for each partition.
Given an input utterance, the $r^{th}$ LM pair provides likelihood probabilities of its behavior lying in the ranges [\textit{1,r+1}] and (\textit{r+1,9}].
We obtain likelihoods from all LM pairs and convert them into a posterior distribution.
Finally, its expected value is output as the behavior score of the input utterance.

For our analysis, we test at all window lengths but only train 3-gram LMs due to the difficulty of training higher-order LMs caused by the \textit{curse of dimensionality} \citep{bengio2003neural}, as explained in \ref{ssec:append_ngram_train}.
We show in Sec.~\ref{ssec:res_winlen} that this mismatch does not particularly bias our results.
Details of the model design and training can be found in \ref{sec:append_ngram_model}.

\subsubsection{Effect of choice of behavior modeling framework}
\label{sssec:other_model}

\begin{figure}[!h]
\centering
\includegraphics[height=0.7\linewidth, width=0.8\linewidth]{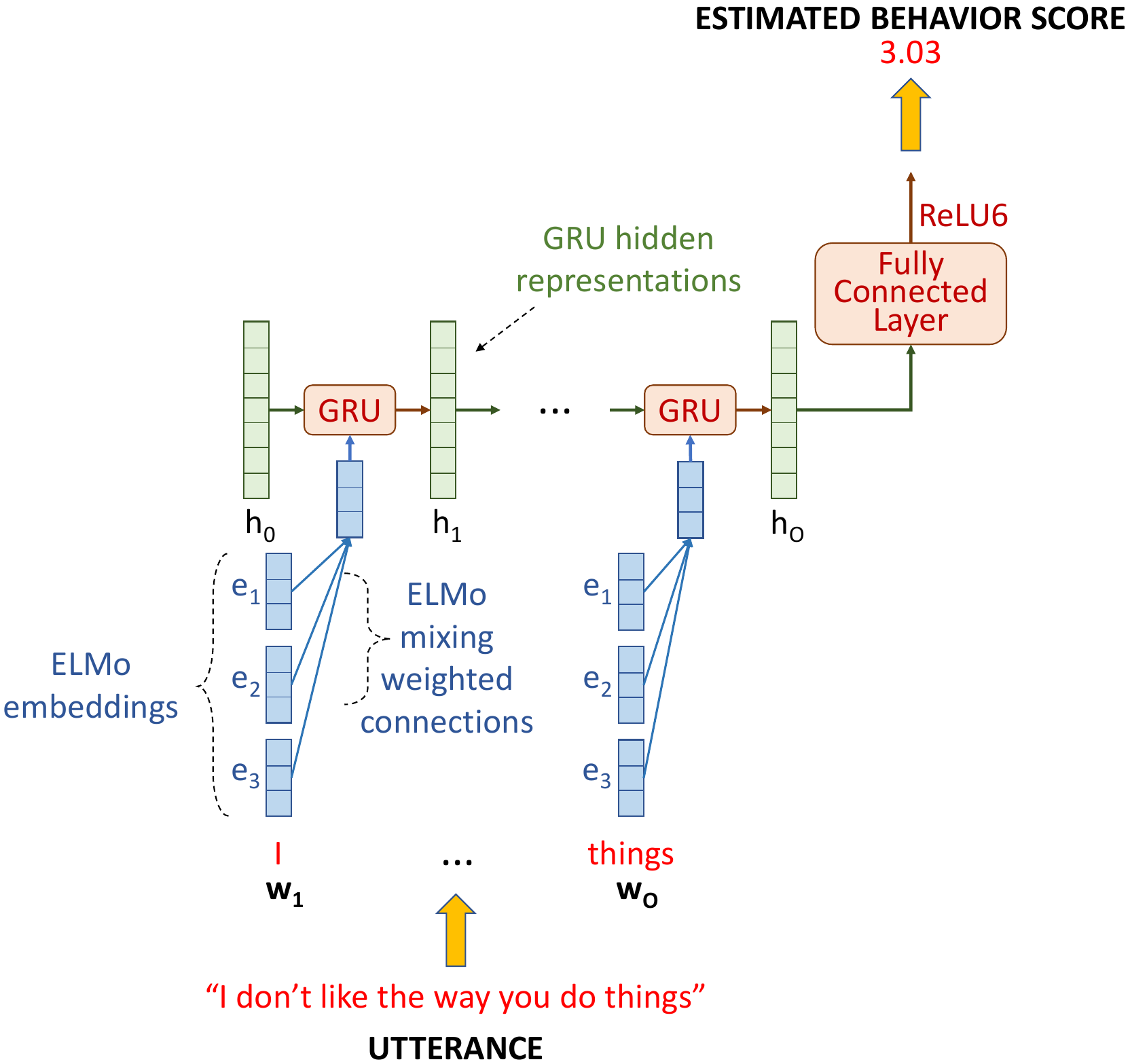}
\caption{\textit{Neural model that uses a $O$-word-long window to estimate the behavior score of a sample utterance:} Given an utterance, the ELMo word embedding sequence is mapped to a fixed-length hidden representation and passed through a fully connected layer and ReLU6 activation to obtain the estimate of the behavior score}
\label{fig:neural_est}
\end{figure}

From Eqn.~\ref{eqn:q} we see that our analysis is fundamentally tied to the behavior model $M$, in which case different choices of $M$ might bias our results differently.
In order to examine this effect, we repeat our analysis with a Recurrent Neural Network model as shown in Figure~\ref{fig:neural_est}.
Since this is only a comparative analysis, we limit its scope by training and testing at two window lengths: short (3 words) and medium-length (30 words).
Efforts to analyze at a long window (100 words) were unsuccessful due to instability during training for behaviors with heavily skewed rating distributions, as explained in \ref{ssec:append_neural_train}.

We use a model similar to the one in \citep{tseng2016couples}, which was shown to accurately estimate \textit{Negative} behavior of speakers in dyadic interactions.
It consists of a Gated Recurrent Unit (GRU) \citep{cho2014properties} followed by a fully connected layer and a \textit{ReLU6} \citep{krizhevsky2010convolutional} activation.
ReLU6 is a modified version of the standard ReLU activation that takes in input $x$ and provides an output $y$ that is bounded between 0 and 6:
\begin{equation}
y = min(max(x, 0), 6)
\end{equation}
Given an input sequence of words, their embeddings are passed to the GRU, which maps them to a fixed-length hidden representation.
This representation is then passed through a fully connected layer and the ReLU6 to obtain the behavior score for that word sequence.

In order to construct the input embeddings, we use ELMo \citep{peters2018deep} representations.
For each word, ELMo provides three embeddings that are mixed using trainable normalized weights to produce a single embedding.
When trained, these weights can be used to gain insights into language use patterns of different behaviors.
Detailed descriptions of our model training and the ELMo mixing weights are provided in \ref{sec:append_neural_model}.

\subsection{Behavioral Grouping}
\label{ssec:beh_groups}

\begin{figure}[!h]
\centering
\includegraphics[height=0.8\linewidth, width=1\linewidth]{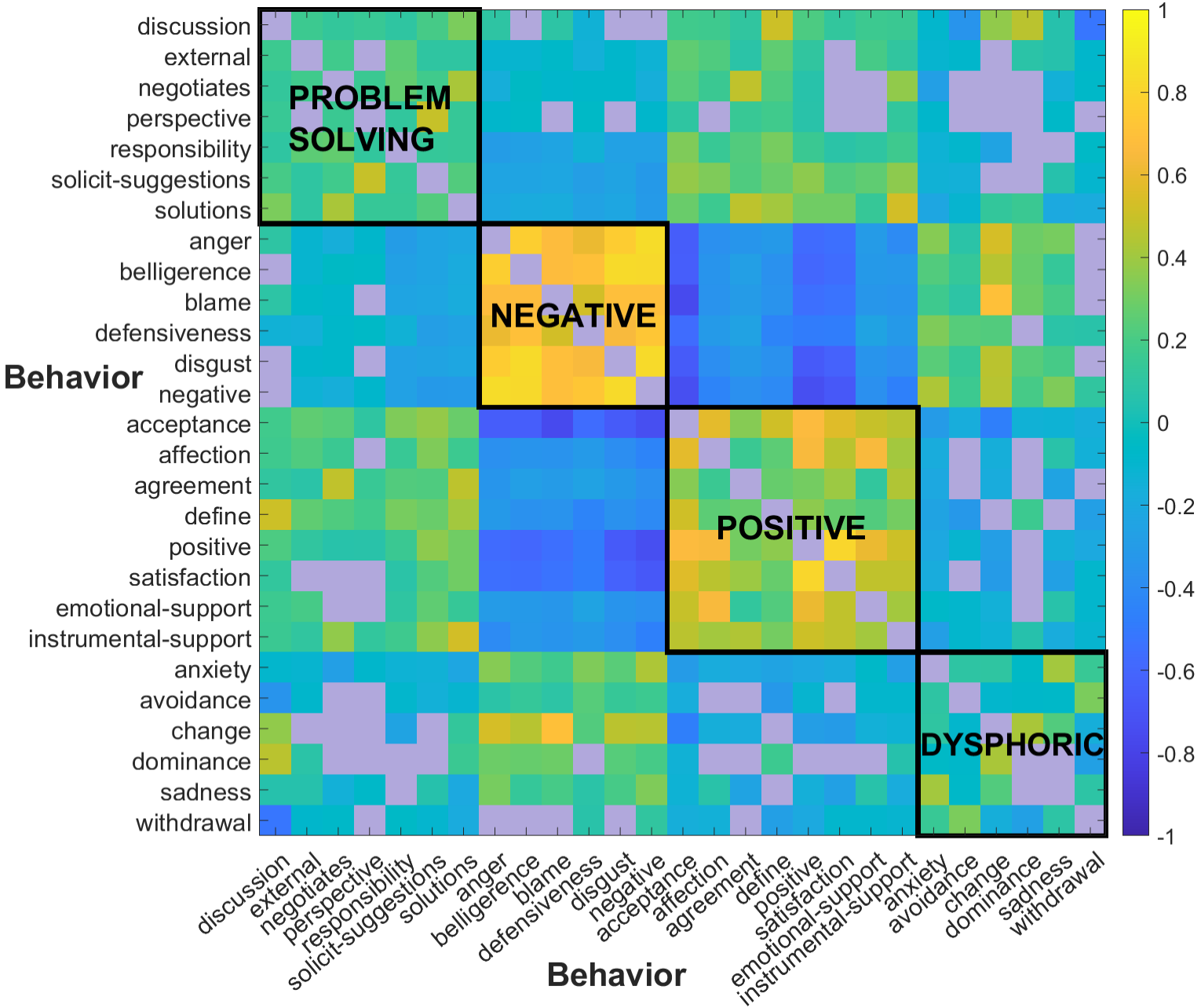}
\caption{\textit{Grouping of Couples Therapy behaviors based on their relation to each other:}
Behaviors are clustered in the space of their human annotations using the k-means algorithm.
The cell in the $(i,j)$ position shows the Spearman Correlation between human ratings of the corresponding behaviors $i$ and $j$.
Yellow (Blue) indicates highly positive (negative) correlation.
Non-diagonal gray cells indicate that correlation is not statistically significant $(p < 0.05)$}
\label{fig:beh_groups}
\end{figure}

As seen in Sec.~\ref{sec:data}, there exist some perceptual relations between the 27 Couples Therapy behaviors based on their definitions.
For example, \textit{Negative} is similar to \textit{Blame} but opposite to \textit{Positive}, while \textit{Withdrawal} is similar to \textit{Avoidance} but opposite to \textit{Discussion}.
Grouping these behaviors based on such relations can lend more interpretability to our analysis and help us better study the link between the nature of a behavior and its observation length.
Hence, we group the 27 behaviors by clustering their human expert ratings using the k-means algorithm described in Algorithm~\ref{alg:beh_group} in \ref{sec:append_alg}.
Figure~\ref{fig:beh_groups} shows the resulting 4 behavior groups.

Our behavior groups closely resemble the work by Sevier et al. \citep{sevier2008observed} which also derived 4 \dq{scales} of behavior using a Principal Component Analysis-based approach: \textit{Negativity}, \textit{Withdrawal}, \textit{Positivity} and \textit{Problem-Solving}.
Hence, we name our 4 groups of behaviors in similar fashion.
The first group is \textbf{Problem-Solving} since it pertains to a back-and-forth style of interaction, with behaviors such as \textit{Discussion}, \textit{Negotiates} and \textit{Solutions}.
The second group consists of behaviors such as \textit{Anger}, \textit{Blame} and \textit{Disgust}; hence, we refer to it as \textbf{Negative}.
Similarly, we name the third group \textbf{Positive} since it contains \textit{Affection}, \textit{Positive}, \textit{Satisfaction}, etc.
The last group contains behaviors such as \textit{Anxiety}, \textit{Sadness} and \textit{Withdrawal}, most of which are related to \dq{dysphoria}, a state of unease or unhappiness; hence, we name this group \textbf{Dysphoric}.

\section{Results \& Discussion}
\label{sec:results}

We now present the results of our analysis experiments.
First, we examine which window lengths are found to be most appropriate for which behaviors and discuss how they relate to the nature and characteristics of the behaviors.
Then, we compare results from different modeling frameworks and comment on patterns that are consistent across models versus patterns that are heavily influenced by the choice of model.
Finally, we consolidate our findings and offer recommendations for the choices of observation window length and modeling framework for different types of behaviors.

In this section, we present only the final results of our analysis framework that show the best window length for behavior estimation in the Ngram and Neural models.
We detail the intermediate BCS results for these models, which show the similarity between aggregate behavior estimates and ground truth ratings, in \ref{ssec:append_res_ngram_bcs} and \ref{ssec:append_res_neural_bcs} respectively.
Similarly, the intermediate BRC results, which show the similarity between inter-behavior relationships in behavior estimates and ground truth ratings, are given in \ref{ssec:append_res_ngram_brc} and \ref{ssec:append_res_neural_brc} respectively.
Based on these, we used thresholds $Y_1 = 0.59$ and $Y_2 = 0.95$ for our analysis procedure.

In general, we observed that both the Ngram model and the Neural model performed best when estimating \textbf{Negative} behaviors with their BCS close to 0.5 on average.
Performance was lower in both models for \textbf{Positive} behaviors, with BCS around 0.4 on average.
In the case of \textbf{Problem-Solving} and \textbf{Dysphoric} behaviors, both models exhibited a wide range of BCS values, from 0.1 to 0.53.
We interpret and compare the performance of both models in greater detail in Sec.~\ref{ssec:res_func}.

\subsection{Relation between observation window length and behavior}
\label{ssec:res_winlen}

\begin{figure}[!h]
\centering
    \includegraphics[height=0.7\linewidth, width=1\linewidth]{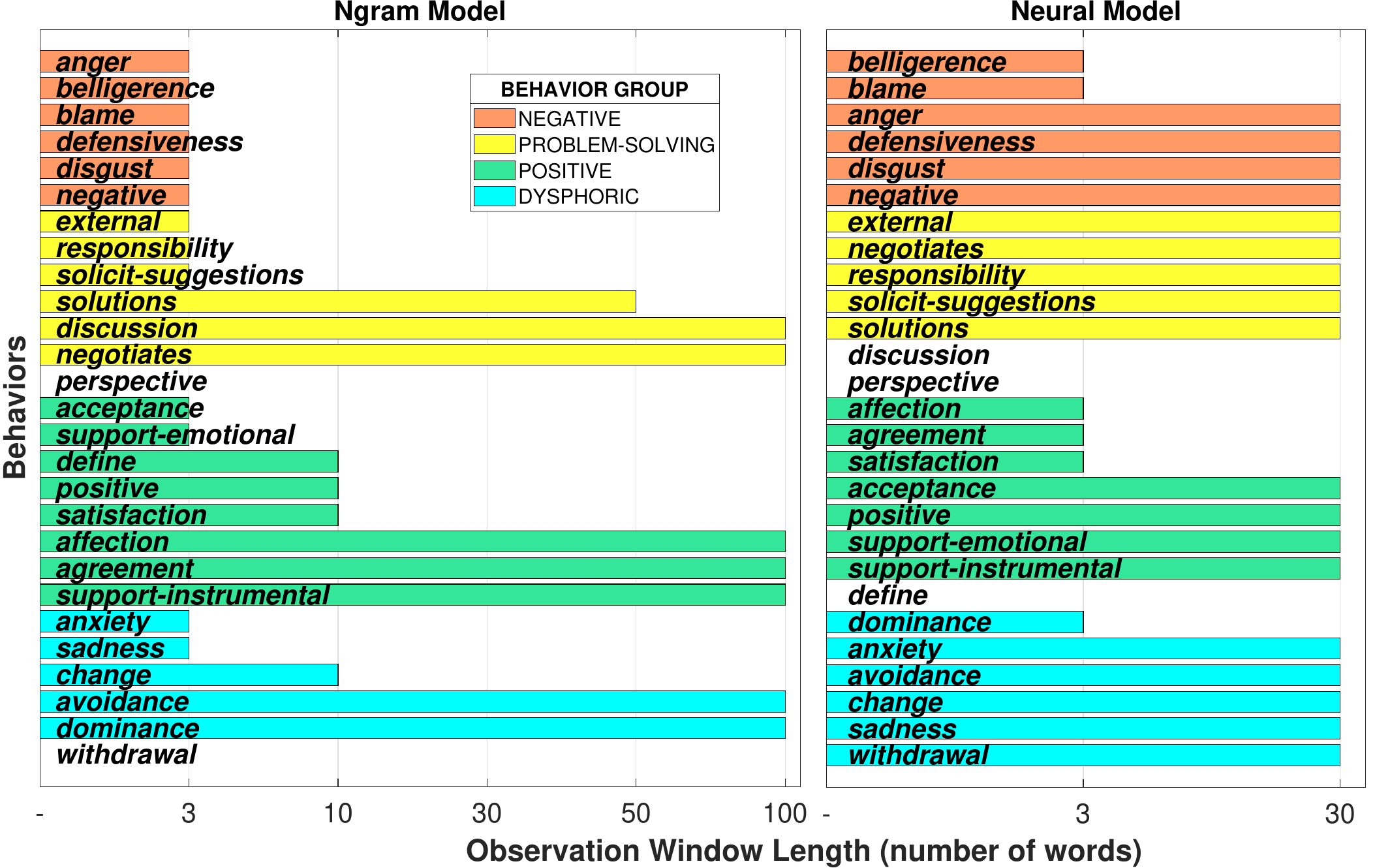}
    \caption{\textit{Appropriate window lengths of behaviors estimated with (left bar plot) N-gram model and (right bar plot) Neural model. Absence of a bar for a behavior implies that our analysis framework was unable to determine appropriate window lengths for that behavior}}
    \label{fig:final_analysis}
\end{figure}

Figure~\ref{fig:final_analysis} shows the final analysis results for both modeling frameworks.
The left bar plot shows the results with the Ngram model at the five window lengths tested: \{\textit{3, 10, 30, 50, 100}\} words.
The right bar plot shows the results with the Neural model at the two window lengths tested: \{\textit{3, 30}\} words.
Each behavior is shown against its appropriate observation window length, as determined by our analysis, and sorted within its behavior group.
We were unable to determine appropriate window lengths for behaviors whose estimates were neither similar nor consistent with human judgments; such behaviors are shown without a bar in the figure.
With the Ngram model, these behaviors are \textit{Perspective} and \textit{Withdrawal} while in the Neural model, they are \textit{Discussion}, \textit{Perspective} and \textit{Define}.

We wish to clarify here that these results should not be interpreted as global optimums.
For example, we see that with the Ngram model, the behavior \textit{Anger} is best estimated using a 3-word long window.
What this means is that 3 words is the best window length, among the ones that we tested, for estimating \textit{Anger}.
Another way to interpret this result is that short windows (3 words) are better than medium-length windows (30 words) or long windows (100 words) for estimating \textit{Anger}.

First, we focus on the results obtained with the Ngram model, shown in the left side plot in Figure~\ref{fig:final_analysis}.
In general, we found that the BCS of behaviors did not vary greatly, as can be seen in Figure~\ref{fig:N-gram_bcs}.
However, in cases where there were statistically significant changes in BCS, we were able to determine appropriate window lengths using Stage 2 of the analysis framework.
For behaviors whose BCS did not significantly change, such as \textit{Affection} and \textit{Solutions}, we were able to determine their window lengths by inspecting their BRC using Stages 3 and 4 of the analysis framework.
We refer the readers to \ref{sec:append_res_ngram} for a more in-depth discussion of the Ngram model analysis results.

We see that most of the behaviors in our dataset tend to perform best with short windows, such as \textit{Acceptance} and \textit{Negative} at 3 words and \textit{Positive} and \textit{Change} at 10 words.
At the same time, we see some behaviors that perform best when scored using much longer observation windows, such as \textit{Solutions} at 50 words and \textit{Avoidance} at 100 words.
We also note that more than half of the behaviors perform better at windows longer than 3 words, even though the N-gram models were trained on 3 words.
This shows that the train-test mismatch mentioned in Sec.~\ref{sssec:mle_model} did not particularly bias our results towards always selecting 3 words as the appropriate window length.

At the group-level, we see that all the \textbf{Negative} behaviors perform best with short observation windows.
This seems to be in line with our intuition about the emotional, short-term nature of these behaviors that lends itself to brief expressions.
These findings match the observation by Baumeister et al. \citep{baumeister2001bad} that humans show heightened awareness of and react more quickly to negative information than to positive information.
They also match the findings by Carney et al. \citep{carney2007thin} who reported that \emph{Negative} affect could be quantified well using thin slices whereas \emph{Positive} affect required thicker slices.

The remaining groups, on the other hand, appear to be expressed over a wide range of lengths.
Among these, \textbf{Positive} and \textbf{Dysphoric} behaviors mostly work best at short observations (10 words or fewer).
For \textbf{Positive} behaviors, this is likely due to their affective content which, while not as brief as negative ones, is nevertheless short-term.
\textbf{Dysphoric} behaviors, on the other hand, are characterized by a lack of participation and expression and are thus, likely to be marked by brief expressions, which could be why they tend to do best at short window lengths.

Finally, \textbf{Problem-Solving} behaviors are evenly split between either very short (3 words) or much longer windows (50 - 100 words).
This is a little surprising since we would normally expect them to be mostly, if not completely, long-range due to their extended, back-and-forth nature.
Upon inspection, we found that these behaviors exhibited their highest similarity with human ratings (as seen from their BCS) at short windows but their highest consistency (as seen from their BRC) at long windows.
This suggests that the existing functionals are unable to effectively aggregate behavior estimates from longer windows.
Hence, for such behaviors, more sophisticated functionals might be able to exploit the full potential of long windows.

Next, we examine the outcome of the comparison analysis with the Neural model and compare them with the above N-gram model results.
This will help us understand how they vary based on the choice of modeling framework.

\subsection{Relation between observation window length and modeling framework}
\label{ssec:res_model}

We now focus on the comparison analysis with the Neural model, shown in the right side plot in Figure~\ref{fig:final_analysis}.
We show the appropriate observation window length for each behavior, sorted by its group.

\begin{figure}[!h]
\centering
    \includegraphics[height=0.5\linewidth, width=1\linewidth]{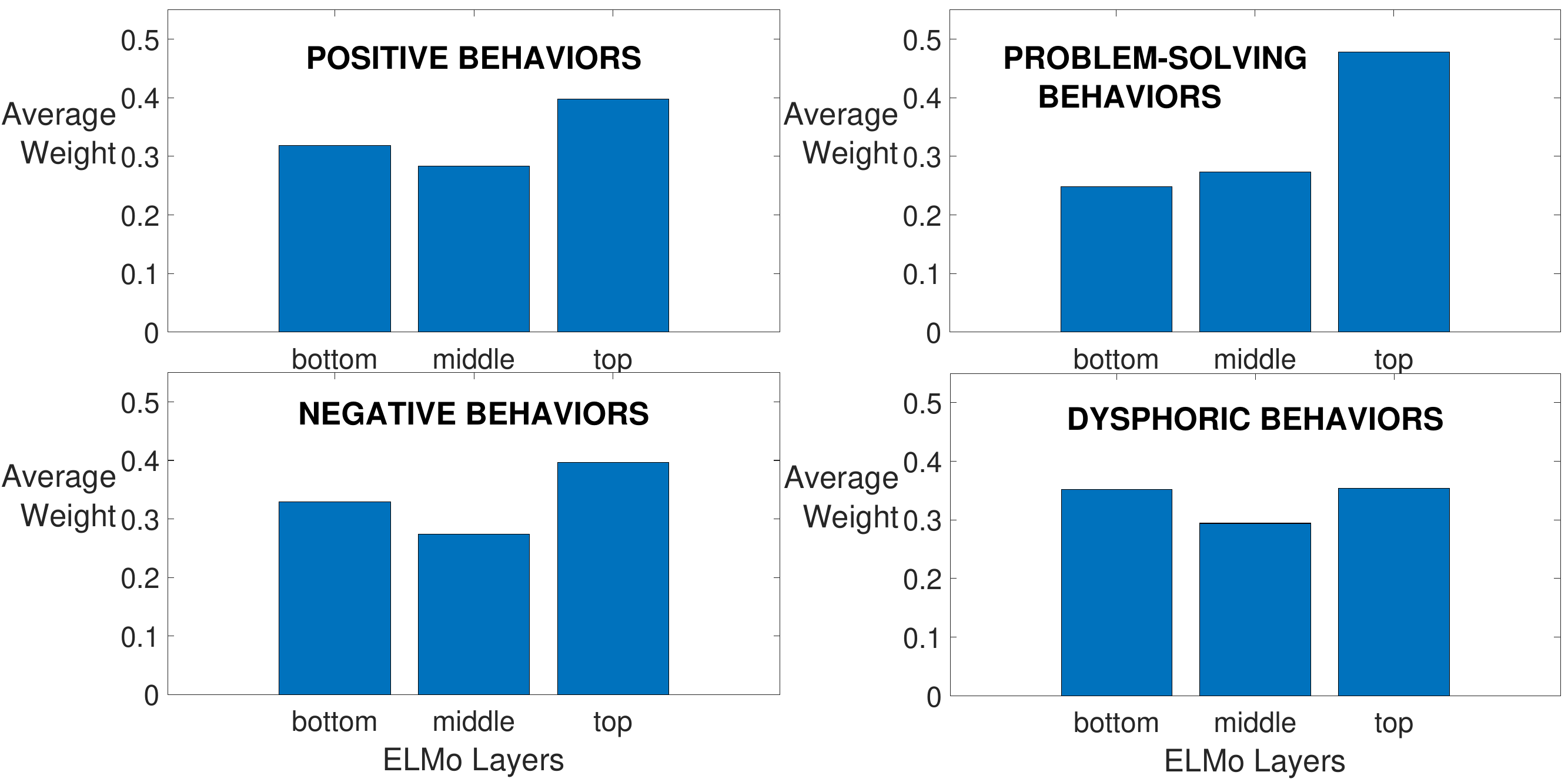}
    \caption{\textit{Trained ELMo layer weights for different behavior groups}}
    \label{fig:neural_weights}
\end{figure}

Once again, we see that \textbf{Negative} and \textbf{Positive} behaviors show a greater preference for short observation windows than \textbf{Problem-Solving} and \textbf{Dysphoric} behaviors.
In particular, all the \textbf{Problem-Solving} behaviors perform best when estimated using longer windows.
This is further supported by the trained ELMo weights in Figure~\ref{fig:neural_weights} which show that \textbf{Problem-Solving} behaviors rely heavily on the top layer, which is associated with complex and high-level language aspects that are typically long-term.
Therefore, the core finding that affect-based behaviors are best captured using shorter window lengths while non-affect-based behaviors are best captured using longer observation windows is seen to hold consistently across models.

There do exist some differences, however, that appear to be driven more by the nature of the modeling framework and less by the behaviors.
For instance, with the Neural model we see that most of the behaviors perform best at 30 words, with only less than a quarter performing well at 3 words.
This is possibly due to its Gated Recurrent Unit, which was originally designed to handle long-context dependencies and, thus, works better when fed information from a longer observation window.

We also see that \textbf{Negative} behaviors perform better at medium-length windows than at short ones, on average.
This is in contrast to the N-gram model, where they all performed best at short window lengths.
To understand the reason for this difference, we inspected their BCS and BRC values and found that while the Neural model's estimates were more consistent at 3 words, they were more similar at 30 words.
This suggests that better functionals might be required to accurately summarize \textbf{Negative} behaviors at short windows when using the Neural model.

\subsection{Relation between behavior and modeling framework}
\label{ssec:res_func}

\begin{figure}[!h]
\centering
    \includegraphics[height=0.5\linewidth, width=1\linewidth]{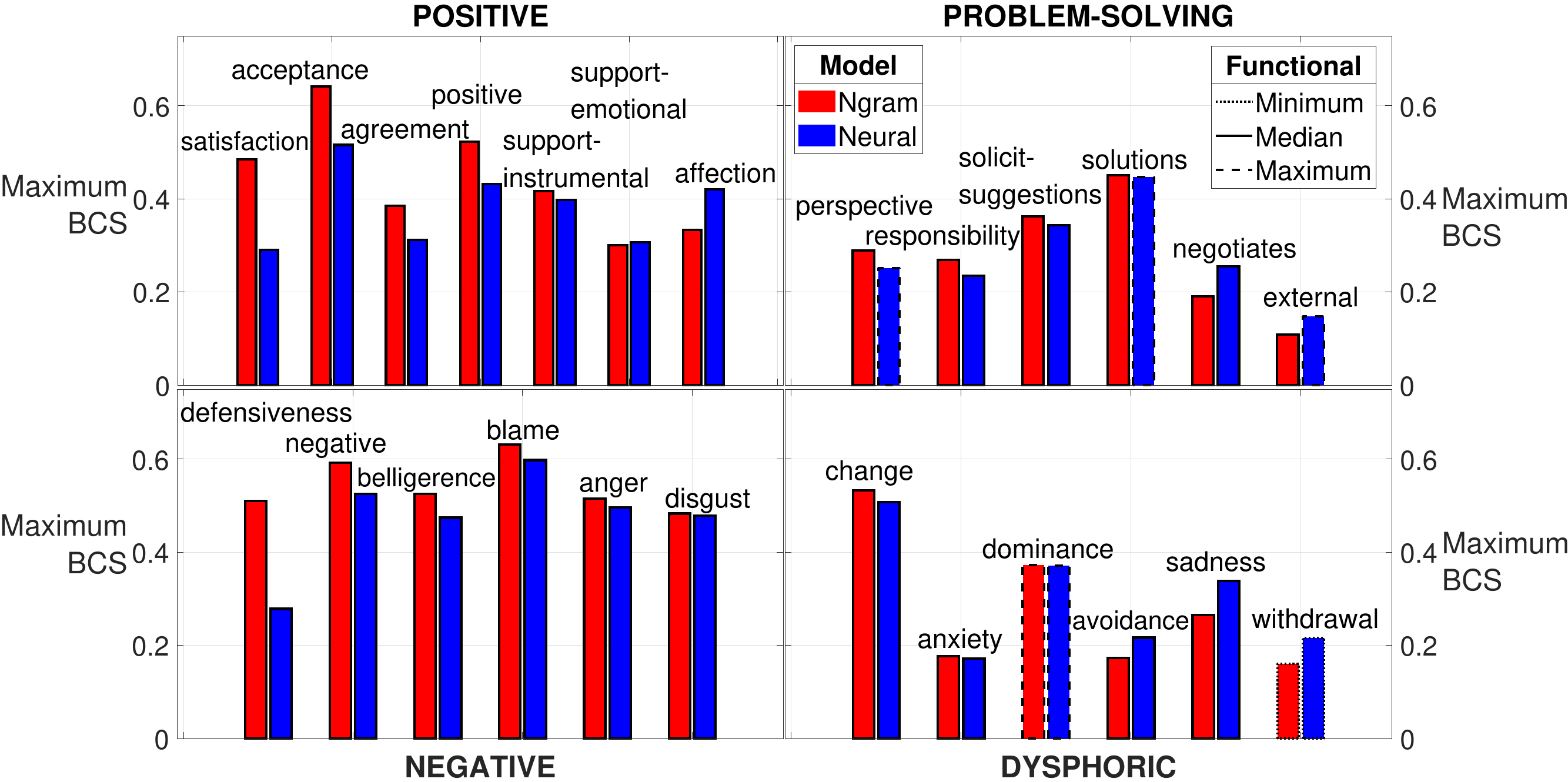}
    \caption{\textit{Comparison of best modeling performance from both models over all window lengths for different behaviors:} Performance here refers to the similarity between behavior estimates and human judgments, as measured by the Behavior Construct Similarity (BCS) metric.}
    \label{fig:model_comparison}
\end{figure}

Finally, we compare the two models in terms of how well they estimate each behavior and which functionals they used in doing so.
This can provide insights into the estimation process and help us understand which of the two models is a better fit for a behavior.
Figure~\ref{fig:model_comparison} shows, for every behavior, the best performance of each model over all window lengths and functionals.

We see that the N-gram model performs as well as, if not better than, the Neural model when quantifying most \textbf{Negative} and \textbf{Positive} behaviors.
While this might seem counter-intuitive, we have observed a similar result in our previous work where N-gram-based and Neural-based models performed similarly when classifying the behavior construct \textit{Negative} \citep{chakravarthula2018modeling}, which is part of the \textbf{Negative} group in this work.
Furthermore, as we saw earlier, most of these behaviors were better quantified at shorter windows than longer ones.
This suggests that short, frequently used expressions carry a considerable amount of information about how much affect a person is expressing.
Since an N-gram model is ideal for capturing fixed-length, short expressions, it appears to be better suited than the Neural model for this task.

In the case of \textbf{Problem-Solving} and \textbf{Dysphoric} behaviors, we see that the Neural model performs as well as, if not better than, the N-gram model.
Since these behaviors are more complex and ambiguous than affect-based ones, estimating them accurately requires the ability to handle long context dependencies in a sophisticated, non-linear manner.
This is precisely the advantage that the Neural model offers over the N-gram model; hence, in line with our expectation, we see that it performs better for these behaviors.

Among the aggregating functionals, we see that the \textit{median} is the best one for all of the \textbf{Negative} and \textbf{Positive} behaviors, similar to previous works \citep{tseng2016couples,tseng2018honey}.
Since the \textit{median} represents the \dq{typical} value, this suggests that affect-based behaviors are steadily expressed throughout the entire interaction, rather than impulsively or rarely.

In the \textbf{Problem-Solving} and \textbf{Dysphoric} groups, however, functionals that represent extreme deviations from the \dq{typical} value are seen to perform well.
In particular, \textit{Dominance}, an overt and high-arousal behavior, is best aggregated as the \textit{maximum} while \textit{Withdrawal}, a subtle and low-arousal behavior, is best aggregated as the \textit{minimum}.
This suggests that the expression patterns of non-affect-based behaviors might be highly infrequent and impulsive.
These findings are thematically aligned with Lee et al. \citep{lee2012based} who showed that humans use different processes for different behaviors when forming an overall impression over the course of an interaction.

In general, while our findings agree with previous works, they diverge slightly from some studies that deal with the audio and video channels.
For instance, while our analysis showed that \textit{Sadness} and \textit{Positive} were best captured at similar window lengths, Krull et al. \citep{krull1998smiles} reported that sad faces evoked less spontaneous reactions than happy faces, implying that they were captured at different window lengths in the visual modality.
Similarly, Li et al. \citep{li2020linking} reported that, in contrast to our findings, behaviors such as \textit{Blame} and \textit{Negativity} performed better with longer observation windows while \textit{Sadness} performed best at shorter windows.

This suggests that affect-based behaviors are sufficiently expressed through all three modalities - audio, video and lexical - but over different time-scales, in which case multi-scale approaches might be beneficial when fusing information across modalities.
Furthermore, dysphoric behaviors such as \textit{Sadness} do not appear to be strongly detected in either the audio or the lexical modality, regardless of how long they are observed, but appear to be well captured in the visual modality.
This provides additional motivation for the use of multimodal approaches when estimating a general set of behaviors.

\section{Conclusions \& Future Work}
\label{sec:conc_futwor}

In this paper, we analyzed how long a system needs to observe conversational language cues, measured in number of words, in order to quantify different behaviors.
We proposed an analysis framework and associated evaluation metrics that can be used to determine appropriate window lengths for behavior estimation, even in scenarios where reference human judgments are not available to compare against at every possible window length.
We applied our analysis to the Couples Therapy dataset which contains a rich and diverse set of behaviors observed in real-life interactions.
We also examined the robustness of our analysis to two different behavior modeling methods, a Maximum Likelihood N-gram model and a Deep Neural Network model.
Finally, we compared our findings with those from similar work in psychology, machine learning and speech processing and addressed pertinent issues related to the nature of human behavior expression in spoken language.

Our analysis showed that affect-based behaviors are steadily and frequently expressed during a conversation and can be reliably captured from short lexical cues.
On the other hand, behaviors involving complex, back-and-forth deliberations tend to be expressed in the form of rare and extreme events and require much longer observation windows in order to be accurately understood.
Finally, the expression of dysphoria appears to be difficult to detect from language alone, even when observed using long windows.

The findings from this work are of relevance not only to machine learning-based behavior estimation approaches but also to psychological research studies that deal with manual annotations of behaviors.
For instance, future studies might find it beneficial, both in terms of cost as well as time, to consider which types of behaviors are of primary importance when deciding on the length of interactions to be collected.
Studies focused on negative and positive affect-based behaviors may be able to elicit and measure them over relatively brief periods of time.
On the other hand, studies focused on discussion-oriented behaviors will likely require considerably longer intervals that can generate larger amounts of text.

The next step in this work would be to extrinsically evaluate our findings across different behavior modeling tasks and checking if they translate into improved performance over using the same window length for all behaviors.
It is also worth investigating how the window length requirements change when employing a multimodal analysis system that uses acoustic and visual cues in addition to the lexical cues.
As a supplement to this work, we would like to crowdsource human annotations of how accurately humans can assess different behaviors using different amounts of text information, thus conducting a study similar to those involving thin slices.
A related effort in that direction would also be to test on datasets with dialog acts and utterance-level annotations of behavior for direct evaluation.
Finally, we plan on investigating if functionals that mimic human-like perception, such as primacy and recency \citep{steiner1989immediate}, might be a better fit for behavior aggregation during an interaction.

\newpage

\section{References}

\bibliographystyle{elsarticle-harv} 
\bibliography{refs}

\begin{thebibliography}{64}
\expandafter\ifx\csname natexlab\endcsname\relax\def\natexlab#1{#1}\fi
\providecommand{\url}[1]{\texttt{#1}}
\providecommand{\href}[2]{#2}
\providecommand{\path}[1]{#1}
\providecommand{\DOIprefix}{doi:}
\providecommand{\ArXivprefix}{arXiv:}
\providecommand{\URLprefix}{URL: }
\providecommand{\Pubmedprefix}{pmid:}
\providecommand{\doi}[1]{\href{http://dx.doi.org/#1}{\path{#1}}}
\providecommand{\Pubmed}[1]{\href{pmid:#1}{\path{#1}}}
\providecommand{\bibinfo}[2]{#2}
\ifx\xfnm\relax \def\xfnm[#1]{\unskip,\space#1}\fi
\bibitem[{Ambady and Rosenthal(1992)}]{ambady1992thin}
\bibinfo{author}{Ambady, N.}, \bibinfo{author}{Rosenthal, R.},
  \bibinfo{year}{1992}.
\newblock \bibinfo{title}{Thin slices of expressive behavior as predictors of
  interpersonal consequences: A meta-analysis.}
\newblock \bibinfo{journal}{Psychological bulletin} \bibinfo{volume}{111},
  \bibinfo{pages}{256}.
\bibitem[{Baer et~al.(2009)Baer, Wells, Rosengren, Hartzler, Beadnell and
  Dunn}]{baer2009agency}
\bibinfo{author}{Baer, J.S.}, \bibinfo{author}{Wells, E.A.},
  \bibinfo{author}{Rosengren, D.B.}, \bibinfo{author}{Hartzler, B.},
  \bibinfo{author}{Beadnell, B.}, \bibinfo{author}{Dunn, C.},
  \bibinfo{year}{2009}.
\newblock \bibinfo{title}{Agency context and tailored training in technology
  transfer: A pilot evaluation of motivational interviewing training for
  community counselors}.
\newblock \bibinfo{journal}{Journal of substance abuse treatment}
  \bibinfo{volume}{37}, \bibinfo{pages}{191--202}.
\bibitem[{Baucom et~al.(2011)Baucom, Sevier, Eldridge, Doss and
  Christensen}]{baucom2011observed}
\bibinfo{author}{Baucom, K.J.}, \bibinfo{author}{Sevier, M.},
  \bibinfo{author}{Eldridge, K.A.}, \bibinfo{author}{Doss, B.D.},
  \bibinfo{author}{Christensen, A.}, \bibinfo{year}{2011}.
\newblock \bibinfo{title}{Observed communication in couples two years after
  integrative and traditional behavioral couple therapy: Outcome and link with
  five-year follow-up.}
\newblock \bibinfo{journal}{Journal of consulting and clinical psychology}
  \bibinfo{volume}{79}, \bibinfo{pages}{565}.
\bibitem[{Baumeister et~al.(2001)Baumeister, Bratslavsky, Finkenauer and
  Vohs}]{baumeister2001bad}
\bibinfo{author}{Baumeister, R.F.}, \bibinfo{author}{Bratslavsky, E.},
  \bibinfo{author}{Finkenauer, C.}, \bibinfo{author}{Vohs, K.D.},
  \bibinfo{year}{2001}.
\newblock \bibinfo{title}{Bad is stronger than good}.
\newblock \bibinfo{journal}{Review of general psychology} \bibinfo{volume}{5},
  \bibinfo{pages}{323--370}.
\bibitem[{Baumeister et~al.(2007)Baumeister, Vohs, Nathan~DeWall and
  Zhang}]{baumeister2007emotion}
\bibinfo{author}{Baumeister, R.F.}, \bibinfo{author}{Vohs, K.D.},
  \bibinfo{author}{Nathan~DeWall, C.}, \bibinfo{author}{Zhang, L.},
  \bibinfo{year}{2007}.
\newblock \bibinfo{title}{How emotion shapes behavior: Feedback, anticipation,
  and reflection, rather than direct causation}.
\newblock \bibinfo{journal}{Personality and social psychology review}
  \bibinfo{volume}{11}, \bibinfo{pages}{167--203}.
\bibitem[{Bengio et~al.(2003)Bengio, Ducharme, Vincent and
  Jauvin}]{bengio2003neural}
\bibinfo{author}{Bengio, Y.}, \bibinfo{author}{Ducharme, R.},
  \bibinfo{author}{Vincent, P.}, \bibinfo{author}{Jauvin, C.},
  \bibinfo{year}{2003}.
\newblock \bibinfo{title}{A neural probabilistic language model}.
\newblock \bibinfo{journal}{Journal of machine learning research}
  \bibinfo{volume}{3}, \bibinfo{pages}{1137--1155}.
\bibitem[{Black et~al.(2013)Black, Katsamanis, Baucom, Lee, Lammert,
  Christensen, Georgiou and Narayanan}]{black2013toward}
\bibinfo{author}{Black, M.P.}, \bibinfo{author}{Katsamanis, A.},
  \bibinfo{author}{Baucom, B.R.}, \bibinfo{author}{Lee, C.C.},
  \bibinfo{author}{Lammert, A.C.}, \bibinfo{author}{Christensen, A.},
  \bibinfo{author}{Georgiou, P.G.}, \bibinfo{author}{Narayanan, S.S.},
  \bibinfo{year}{2013}.
\newblock \bibinfo{title}{Toward automating a human behavioral coding system
  for married couples’ interactions using speech acoustic features}.
\newblock \bibinfo{journal}{Speech communication} \bibinfo{volume}{55},
  \bibinfo{pages}{1--21}.
\bibitem[{Blackman and Funder(1998)}]{blackman1998effect}
\bibinfo{author}{Blackman, M.C.}, \bibinfo{author}{Funder, D.C.},
  \bibinfo{year}{1998}.
\newblock \bibinfo{title}{The effect of information on consensus and accuracy
  in personality judgment}.
\newblock \bibinfo{journal}{Journal of Experimental Social Psychology}
  \bibinfo{volume}{34}, \bibinfo{pages}{164--181}.
\bibitem[{Busso et~al.(2008)Busso, Bulut, Lee, Kazemzadeh, Mower, Kim, Chang,
  Lee and Narayanan}]{busso2008iemocap}
\bibinfo{author}{Busso, C.}, \bibinfo{author}{Bulut, M.}, \bibinfo{author}{Lee,
  C.C.}, \bibinfo{author}{Kazemzadeh, A.}, \bibinfo{author}{Mower, E.},
  \bibinfo{author}{Kim, S.}, \bibinfo{author}{Chang, J.N.},
  \bibinfo{author}{Lee, S.}, \bibinfo{author}{Narayanan, S.S.},
  \bibinfo{year}{2008}.
\newblock \bibinfo{title}{Iemocap: Interactive emotional dyadic motion capture
  database}.
\newblock \bibinfo{journal}{Language resources and evaluation}
  \bibinfo{volume}{42}, \bibinfo{pages}{335}.
\bibitem[{Carney et~al.(2007)Carney, Colvin and Hall}]{carney2007thin}
\bibinfo{author}{Carney, D.R.}, \bibinfo{author}{Colvin, C.R.},
  \bibinfo{author}{Hall, J.A.}, \bibinfo{year}{2007}.
\newblock \bibinfo{title}{A thin slice perspective on the accuracy of first
  impressions}.
\newblock \bibinfo{journal}{Journal of Research in Personality}
  \bibinfo{volume}{41}, \bibinfo{pages}{1054--1072}.
\bibitem[{Chakravarthula et~al.(2018)Chakravarthula, Baucom and
  Georgiou}]{chakravarthula2018modeling}
\bibinfo{author}{Chakravarthula, S.N.}, \bibinfo{author}{Baucom, B.},
  \bibinfo{author}{Georgiou, P.}, \bibinfo{year}{2018}.
\newblock \bibinfo{title}{Modeling interpersonal influence of verbal behavior
  in couples therapy dyadic interactions}.
\newblock \bibinfo{journal}{Proc. Interspeech 2018} ,
  \bibinfo{pages}{2339--2343}.
\bibitem[{Chakravarthula et~al.(2015a)Chakravarthula, Gupta, Baucom and
  Georgiou}]{chakravarthula2015language}
\bibinfo{author}{Chakravarthula, S.N.}, \bibinfo{author}{Gupta, R.},
  \bibinfo{author}{Baucom, B.}, \bibinfo{author}{Georgiou, P.},
  \bibinfo{year}{2015}a.
\newblock \bibinfo{title}{A language-based generative model framework for
  behavioral analysis of couples' therapy}, in: \bibinfo{booktitle}{Acoustics,
  Speech and Signal Processing (ICASSP), 2015 IEEE International Conference
  on}, \bibinfo{organization}{IEEE}. pp. \bibinfo{pages}{2090--2094}.
\bibitem[{Chakravarthula et~al.(2015b)Chakravarthula, Xiao, Imel, Atkins and
  Georgiou}]{chakravarthula2015assessing}
\bibinfo{author}{Chakravarthula, S.N.}, \bibinfo{author}{Xiao, B.},
  \bibinfo{author}{Imel, Z.E.}, \bibinfo{author}{Atkins, D.C.},
  \bibinfo{author}{Georgiou, P.G.}, \bibinfo{year}{2015}b.
\newblock \bibinfo{title}{Assessing empathy using static and dynamic behavior
  models based on therapist's language in addiction counseling}, in:
  \bibinfo{booktitle}{Sixteenth Annual Conference of the International Speech
  Communication Association}.
\bibitem[{Cho et~al.(2014)Cho, van Merrienboer, Bahdanau and
  Bengio}]{cho2014properties}
\bibinfo{author}{Cho, K.}, \bibinfo{author}{van Merrienboer, B.},
  \bibinfo{author}{Bahdanau, D.}, \bibinfo{author}{Bengio, Y.},
  \bibinfo{year}{2014}.
\newblock \bibinfo{title}{On the properties of neural machine translation:
  Encoder--decoder approaches}, in: \bibinfo{booktitle}{Proceedings of SSST-8,
  Eighth Workshop on Syntax, Semantics and Structure in Statistical
  Translation}, pp. \bibinfo{pages}{103--111}.
\bibitem[{Christensen et~al.(2004)Christensen, Atkins, Berns, Wheeler, Baucom
  and Simpson}]{christensen2004}
\bibinfo{author}{Christensen, A.}, \bibinfo{author}{Atkins, D.},
  \bibinfo{author}{Berns, S.}, \bibinfo{author}{Wheeler, J.},
  \bibinfo{author}{Baucom, D.}, \bibinfo{author}{Simpson, L.},
  \bibinfo{year}{2004}.
\newblock \bibinfo{title}{{Traditional versus integrative behavioral couple
  therapy for significantly and chronically distressed married couples}}.
\newblock \bibinfo{journal}{Journal of Consulting and Clinical Psychology}
  \bibinfo{volume}{72}, \bibinfo{pages}{176--191}.
\bibitem[{Cullen and Harte(2017)}]{cullen2017thin}
\bibinfo{author}{Cullen, A.}, \bibinfo{author}{Harte, N.},
  \bibinfo{year}{2017}.
\newblock \bibinfo{title}{Thin slicing to predict viewer impressions of ted
  talks}, in: \bibinfo{booktitle}{Proceedings of the 14th International
  Conference on Auditory-Visual Speech Processing}.
\bibitem[{Diedenhofen and Musch(2015)}]{diedenhofen2015cocor}
\bibinfo{author}{Diedenhofen, B.}, \bibinfo{author}{Musch, J.},
  \bibinfo{year}{2015}.
\newblock \bibinfo{title}{cocor: A comprehensive solution for the statistical
  comparison of correlations}.
\newblock \bibinfo{journal}{PloS one} \bibinfo{volume}{10},
  \bibinfo{pages}{e0121945}.
\bibitem[{Frank and Hall(2001)}]{frank2001simple}
\bibinfo{author}{Frank, E.}, \bibinfo{author}{Hall, M.}, \bibinfo{year}{2001}.
\newblock \bibinfo{title}{A simple approach to ordinal classification}, in:
  \bibinfo{booktitle}{European Conference on Machine Learning},
  \bibinfo{organization}{Springer}. pp. \bibinfo{pages}{145--156}.
\bibitem[{Georgiou et~al.(2011)Georgiou, Black, Lammert, Baucom and
  Narayanan}]{georgiou2011s}
\bibinfo{author}{Georgiou, P.G.}, \bibinfo{author}{Black, M.P.},
  \bibinfo{author}{Lammert, A.C.}, \bibinfo{author}{Baucom, B.R.},
  \bibinfo{author}{Narayanan, S.S.}, \bibinfo{year}{2011}.
\newblock \bibinfo{title}{“that’s aggravating, very aggravating”: Is it
  possible to classify behaviors in couple interactions using automatically
  derived lexical features?}, in: \bibinfo{booktitle}{International Conference
  on Affective Computing and Intelligent Interaction},
  \bibinfo{organization}{Springer}. pp. \bibinfo{pages}{87--96}.
\bibitem[{Gibson et~al.(2016)Gibson, Can, Xiao, Imel, Atkins, Georgiou and
  Narayanan}]{gibson2016deep}
\bibinfo{author}{Gibson, J.}, \bibinfo{author}{Can, D.}, \bibinfo{author}{Xiao,
  B.}, \bibinfo{author}{Imel, Z.E.}, \bibinfo{author}{Atkins, D.C.},
  \bibinfo{author}{Georgiou, P.}, \bibinfo{author}{Narayanan, S.},
  \bibinfo{year}{2016}.
\newblock \bibinfo{title}{A deep learning approach to modeling empathy in
  addiction counseling}.
\newblock \bibinfo{journal}{Commitment} \bibinfo{volume}{111},
  \bibinfo{pages}{21}.
\bibitem[{Graves(2013)}]{graves2013generating}
\bibinfo{author}{Graves, A.}, \bibinfo{year}{2013}.
\newblock \bibinfo{title}{Generating sequences with recurrent neural networks}.
\newblock \bibinfo{journal}{arXiv preprint arXiv:1308.0850} .
\bibitem[{Gupta et~al.(2014)Gupta, Malandrakis, Xiao, Guha, Van~Segbroeck,
  Black, Potamianos and Narayanan}]{gupta2014multimodal}
\bibinfo{author}{Gupta, R.}, \bibinfo{author}{Malandrakis, N.},
  \bibinfo{author}{Xiao, B.}, \bibinfo{author}{Guha, T.},
  \bibinfo{author}{Van~Segbroeck, M.}, \bibinfo{author}{Black, M.},
  \bibinfo{author}{Potamianos, A.}, \bibinfo{author}{Narayanan, S.},
  \bibinfo{year}{2014}.
\newblock \bibinfo{title}{Multimodal prediction of affective dimensions and
  depression in human-computer interactions}, in:
  \bibinfo{booktitle}{Proceedings of the 4th International Workshop on
  Audio/Visual Emotion Challenge}, \bibinfo{organization}{ACM}. pp.
  \bibinfo{pages}{33--40}.
\bibitem[{Heavey et~al.(2002)Heavey, Gill and Christensen}]{heavey2002couples}
\bibinfo{author}{Heavey, C.}, \bibinfo{author}{Gill, D.},
  \bibinfo{author}{Christensen, A.}, \bibinfo{year}{2002}.
\newblock \bibinfo{title}{Couples interaction rating system 2 (cirs2)}.
\newblock \bibinfo{journal}{University of California, Los Angeles}
  \bibinfo{volume}{7}.
\bibitem[{Heyman(2004)}]{heyman2004rapid}
\bibinfo{author}{Heyman, R.E.}, \bibinfo{year}{2004}.
\newblock \bibinfo{title}{Rapid marital interaction coding system (rmics)}, in:
  \bibinfo{booktitle}{Couple observational coding systems}.
  \bibinfo{publisher}{Routledge}, pp. \bibinfo{pages}{81--108}.
\bibitem[{Hochreiter and Schmidhuber(1997)}]{hochreiter1997long}
\bibinfo{author}{Hochreiter, S.}, \bibinfo{author}{Schmidhuber, J.},
  \bibinfo{year}{1997}.
\newblock \bibinfo{title}{Long short-term memory}.
\newblock \bibinfo{journal}{Neural computation} \bibinfo{volume}{9},
  \bibinfo{pages}{1735--1780}.
\bibitem[{Huang et~al.(2017)Huang, Li, Pleiss, Liu, Hopcroft and
  Weinberger}]{huang2017snapshot}
\bibinfo{author}{Huang, G.}, \bibinfo{author}{Li, Y.}, \bibinfo{author}{Pleiss,
  G.}, \bibinfo{author}{Liu, Z.}, \bibinfo{author}{Hopcroft, J.E.},
  \bibinfo{author}{Weinberger, K.Q.}, \bibinfo{year}{2017}.
\newblock \bibinfo{title}{Snapshot ensembles: Train 1, get m for free}, in:
  \bibinfo{booktitle}{International Conference on Learning Representations}.
\bibitem[{Jawahar et~al.(2019)Jawahar, Sagot, Seddah, Unicomb, I{\~n}iguez,
  Karsai, L{\'e}o, Karsai, Sarraute, Fleury et~al.}]{jawahar2019does}
\bibinfo{author}{Jawahar, G.}, \bibinfo{author}{Sagot, B.},
  \bibinfo{author}{Seddah, D.}, \bibinfo{author}{Unicomb, S.},
  \bibinfo{author}{I{\~n}iguez, G.}, \bibinfo{author}{Karsai, M.},
  \bibinfo{author}{L{\'e}o, Y.}, \bibinfo{author}{Karsai, M.},
  \bibinfo{author}{Sarraute, C.}, \bibinfo{author}{Fleury, {\'E}.}, et~al.,
  \bibinfo{year}{2019}.
\newblock \bibinfo{title}{What does bert learn about the structure of
  language?}, in: \bibinfo{booktitle}{57th Annual Meeting of the Association
  for Computational Linguistics (ACL), Florence, Italy}.
\bibitem[{Jones and Christensen(1998)}]{jones1998couples}
\bibinfo{author}{Jones, J.}, \bibinfo{author}{Christensen, A.},
  \bibinfo{year}{1998}.
\newblock \bibinfo{title}{Couples interaction study: Social support interaction
  rating system}.
\newblock \bibinfo{journal}{University of California, Los Angeles}
  \bibinfo{volume}{7}.
\bibitem[{Kingma and Ba(2015)}]{kingma2014adam}
\bibinfo{author}{Kingma, D.P.}, \bibinfo{author}{Ba, J.}, \bibinfo{year}{2015}.
\newblock \bibinfo{title}{Adam: A method for stochastic optimization}, in:
  \bibinfo{booktitle}{International Conference on Learning Representations}.
\bibitem[{Krizhevsky and Hinton(2010)}]{krizhevsky2010convolutional}
\bibinfo{author}{Krizhevsky, A.}, \bibinfo{author}{Hinton, G.},
  \bibinfo{year}{2010}.
\newblock \bibinfo{title}{Convolutional deep belief networks on cifar-10}.
\newblock \bibinfo{journal}{Unpublished manuscript} \bibinfo{volume}{40},
  \bibinfo{pages}{1--9}.
\bibitem[{Krull and Dil(1998)}]{krull1998smiles}
\bibinfo{author}{Krull, D.S.}, \bibinfo{author}{Dil, J.C.},
  \bibinfo{year}{1998}.
\newblock \bibinfo{title}{Do smiles elicit more inferences than do frowns? the
  effect of emotional valence on the production of spontaneous inferences}.
\newblock \bibinfo{journal}{Personality and Social Psychology Bulletin}
  \bibinfo{volume}{24}, \bibinfo{pages}{289--300}.
\bibitem[{Krzyzaniak et~al.(2019)Krzyzaniak, Colman, Letzring, McDonald and
  Biesanz}]{krzyzaniakeffect}
\bibinfo{author}{Krzyzaniak, S.L.}, \bibinfo{author}{Colman, D.E.},
  \bibinfo{author}{Letzring, T.D.}, \bibinfo{author}{McDonald, J.S.},
  \bibinfo{author}{Biesanz, J.C.}, \bibinfo{year}{2019}.
\newblock \bibinfo{title}{The effect of information quantity on distinctive
  accuracy and normativity of personality trait judgments}.
\newblock \bibinfo{journal}{European Journal of Personality} .
\bibitem[{Lee et~al.(2010)Lee, Black, Katsamanis, Lammert, Baucom, Christensen,
  Georgiou and Narayanan}]{lee2010quantification}
\bibinfo{author}{Lee, C.C.}, \bibinfo{author}{Black, M.},
  \bibinfo{author}{Katsamanis, A.}, \bibinfo{author}{Lammert, A.C.},
  \bibinfo{author}{Baucom, B.R.}, \bibinfo{author}{Christensen, A.},
  \bibinfo{author}{Georgiou, P.G.}, \bibinfo{author}{Narayanan, S.S.},
  \bibinfo{year}{2010}.
\newblock \bibinfo{title}{Quantification of prosodic entrainment in affective
  spontaneous spoken interactions of married couples}, in:
  \bibinfo{booktitle}{Eleventh Annual Conference of the International Speech
  Communication Association}.
\bibitem[{Lee et~al.(2014)Lee, Katsamanis, Black, Baucom, Christensen, Georgiou
  and Narayanan}]{lee2014computing}
\bibinfo{author}{Lee, C.C.}, \bibinfo{author}{Katsamanis, A.},
  \bibinfo{author}{Black, M.P.}, \bibinfo{author}{Baucom, B.R.},
  \bibinfo{author}{Christensen, A.}, \bibinfo{author}{Georgiou, P.G.},
  \bibinfo{author}{Narayanan, S.S.}, \bibinfo{year}{2014}.
\newblock \bibinfo{title}{Computing vocal entrainment: A signal-derived
  pca-based quantification scheme with application to affect analysis in
  married couple interactions}.
\newblock \bibinfo{journal}{Computer Speech \& Language} \bibinfo{volume}{28},
  \bibinfo{pages}{518--539}.
\bibitem[{Lee et~al.(2012)Lee, Katsamanis, Georgiou and
  Narayanan}]{lee2012based}
\bibinfo{author}{Lee, C.C.}, \bibinfo{author}{Katsamanis, A.},
  \bibinfo{author}{Georgiou, P.G.}, \bibinfo{author}{Narayanan, S.S.},
  \bibinfo{year}{2012}.
\newblock \bibinfo{title}{Based on isolated saliency or causal integration?
  toward a better understanding of human annotation process using multiple
  instance learning and sequential probability ratio test}, in:
  \bibinfo{booktitle}{Thirteenth Annual Conference of the International Speech
  Communication Association}.
\bibitem[{Li et~al.(2020)Li, Baucom and Georgiou}]{li2020linking}
\bibinfo{author}{Li, H.}, \bibinfo{author}{Baucom, B.},
  \bibinfo{author}{Georgiou, P.}, \bibinfo{year}{2020}.
\newblock \bibinfo{title}{Linking emotions to behaviors through deep transfer
  learning}.
\newblock \bibinfo{journal}{PeerJ Computer Science} \bibinfo{volume}{6},
  \bibinfo{pages}{e246}.
\bibitem[{McCrae et~al.(1986)McCrae, Costa~Jr and Busch}]{mccrae1986evaluating}
\bibinfo{author}{McCrae, R.R.}, \bibinfo{author}{Costa~Jr, P.T.},
  \bibinfo{author}{Busch, C.M.}, \bibinfo{year}{1986}.
\newblock \bibinfo{title}{Evaluating comprehensiveness in personality systems:
  The california q-set and the five-factor model}.
\newblock \bibinfo{journal}{Journal of Personality} \bibinfo{volume}{54},
  \bibinfo{pages}{430--446}.
\bibitem[{Mikolov et~al.(2013)Mikolov, Sutskever, Chen, Corrado and
  Dean}]{mikolov2013distributed}
\bibinfo{author}{Mikolov, T.}, \bibinfo{author}{Sutskever, I.},
  \bibinfo{author}{Chen, K.}, \bibinfo{author}{Corrado, G.S.},
  \bibinfo{author}{Dean, J.}, \bibinfo{year}{2013}.
\newblock \bibinfo{title}{Distributed representations of words and phrases and
  their compositionality}, in: \bibinfo{booktitle}{Advances in neural
  information processing systems}, pp. \bibinfo{pages}{3111--3119}.
\bibitem[{Morales et~al.(2018)Morales, Scherer and
  Levitan}]{morales2018linguistically}
\bibinfo{author}{Morales, M.R.}, \bibinfo{author}{Scherer, S.},
  \bibinfo{author}{Levitan, R.}, \bibinfo{year}{2018}.
\newblock \bibinfo{title}{A linguistically-informed fusion approach for
  multimodal depression detection}, in: \bibinfo{booktitle}{NAACL HLT},
  p.~\bibinfo{pages}{13}.
\bibitem[{Moyers et~al.(2003)Moyers, Martin, Manuel, Miller and
  Ernst}]{moyers2003motivational}
\bibinfo{author}{Moyers, T.B.}, \bibinfo{author}{Martin, T.},
  \bibinfo{author}{Manuel, J.K.}, \bibinfo{author}{Miller, W.R.},
  \bibinfo{author}{Ernst, D.}, \bibinfo{year}{2003}.
\newblock \bibinfo{title}{The motivational interviewing treatment integrity
  (miti) code: Version 2.0}.
\newblock \bibinfo{journal}{Retrieved from Verf{\"u}bar unter: www. casaa. unm.
  edu [01.03. 2005]} .
\bibitem[{Murphy et~al.(2018)Murphy, Hall, Ruben, Frauendorfer, Schmid~Mast,
  Johnson and Nguyen}]{murphy2018predictive}
\bibinfo{author}{Murphy, N.A.}, \bibinfo{author}{Hall, J.A.},
  \bibinfo{author}{Ruben, M.A.}, \bibinfo{author}{Frauendorfer, D.},
  \bibinfo{author}{Schmid~Mast, M.}, \bibinfo{author}{Johnson, K.E.},
  \bibinfo{author}{Nguyen, L.}, \bibinfo{year}{2018}.
\newblock \bibinfo{title}{Predictive validity of thin-slice nonverbal behavior
  from social interactions}.
\newblock \bibinfo{journal}{Personality and Social Psychology Bulletin} ,
  \bibinfo{pages}{0146167218802834}.
\bibitem[{Narayanan and Georgiou(2013)}]{narayanan2013behavioral}
\bibinfo{author}{Narayanan, S.}, \bibinfo{author}{Georgiou, P.G.},
  \bibinfo{year}{2013}.
\newblock \bibinfo{title}{Behavioral signal processing: Deriving human
  behavioral informatics from speech and language}.
\newblock \bibinfo{journal}{Proceedings of the IEEE} \bibinfo{volume}{101},
  \bibinfo{pages}{1203--1233}.
\bibitem[{Nolan(2003)}]{nolan2003stable}
\bibinfo{author}{Nolan, J.}, \bibinfo{year}{2003}.
\newblock \bibinfo{title}{Stable distributions: models for heavy-tailed data}.
\bibitem[{{\"O}hman et~al.(2001){\"O}hman, Lundqvist and
  Esteves}]{ohman2001face}
\bibinfo{author}{{\"O}hman, A.}, \bibinfo{author}{Lundqvist, D.},
  \bibinfo{author}{Esteves, F.}, \bibinfo{year}{2001}.
\newblock \bibinfo{title}{The face in the crowd revisited: a threat advantage
  with schematic stimuli.}
\newblock \bibinfo{journal}{Journal of personality and social psychology}
  \bibinfo{volume}{80}, \bibinfo{pages}{381}.
\bibitem[{Olah et~al.(2017)Olah, Mordvintsev and Schubert}]{olah2017feature}
\bibinfo{author}{Olah, C.}, \bibinfo{author}{Mordvintsev, A.},
  \bibinfo{author}{Schubert, L.}, \bibinfo{year}{2017}.
\newblock \bibinfo{title}{Feature visualization}.
\newblock \bibinfo{journal}{Distill} \bibinfo{volume}{2}, \bibinfo{pages}{e7}.
\bibitem[{P{\'e}rez-Rosas et~al.(2017)P{\'e}rez-Rosas, Mihalcea, Resnicow,
  Singh and An}]{perez2017understanding}
\bibinfo{author}{P{\'e}rez-Rosas, V.}, \bibinfo{author}{Mihalcea, R.},
  \bibinfo{author}{Resnicow, K.}, \bibinfo{author}{Singh, S.},
  \bibinfo{author}{An, L.}, \bibinfo{year}{2017}.
\newblock \bibinfo{title}{Understanding and predicting empathic behavior in
  counseling therapy}, in: \bibinfo{booktitle}{Proceedings of the 55th Annual
  Meeting of the Association for Computational Linguistics (Volume 1: Long
  Papers)}, pp. \bibinfo{pages}{1426--1435}.
\bibitem[{Peters et~al.(2018)Peters, Neumann, Iyyer, Gardner, Clark, Lee and
  Zettlemoyer}]{peters2018deep}
\bibinfo{author}{Peters, M.}, \bibinfo{author}{Neumann, M.},
  \bibinfo{author}{Iyyer, M.}, \bibinfo{author}{Gardner, M.},
  \bibinfo{author}{Clark, C.}, \bibinfo{author}{Lee, K.},
  \bibinfo{author}{Zettlemoyer, L.}, \bibinfo{year}{2018}.
\newblock \bibinfo{title}{Deep contextualized word representations}, in:
  \bibinfo{booktitle}{Proceedings of the 2018 Conference of the North American
  Chapter of the Association for Computational Linguistics: Human Language
  Technologies, Volume 1 (Long Papers)}, pp. \bibinfo{pages}{2227--2237}.
\bibitem[{Reblin et~al.(2019)Reblin, Baucom, Clayton, Utz, Caserta, Lund,
  Mooney and Ellington}]{reblin2019communication}
\bibinfo{author}{Reblin, M.}, \bibinfo{author}{Baucom, B.R.},
  \bibinfo{author}{Clayton, M.F.}, \bibinfo{author}{Utz, R.},
  \bibinfo{author}{Caserta, M.}, \bibinfo{author}{Lund, D.},
  \bibinfo{author}{Mooney, K.}, \bibinfo{author}{Ellington, L.},
  \bibinfo{year}{2019}.
\newblock \bibinfo{title}{Communication of emotion in home hospice cancer care:
  Implications for spouse caregiver depression into bereavement}.
\newblock \bibinfo{journal}{Psycho-Oncology} .
\bibitem[{Rozgi{\'c} et~al.(2011)Rozgi{\'c}, Xiao, Katsamanis, Baucom, Georgiou
  and Narayanan}]{rozgic2011estimation}
\bibinfo{author}{Rozgi{\'c}, V.}, \bibinfo{author}{Xiao, B.},
  \bibinfo{author}{Katsamanis, A.}, \bibinfo{author}{Baucom, B.},
  \bibinfo{author}{Georgiou, P.G.}, \bibinfo{author}{Narayanan, S.},
  \bibinfo{year}{2011}.
\newblock \bibinfo{title}{Estimation of ordinal approach-avoidance labels in
  dyadic interactions: Ordinal logistic regression approach}, in:
  \bibinfo{booktitle}{2011 IEEE International Conference on Acoustics, Speech
  and Signal Processing (ICASSP)}, \bibinfo{organization}{IEEE}. pp.
  \bibinfo{pages}{2368--2371}.
\bibitem[{Satterstrom et~al.(2019)Satterstrom, Polzer, Kwan, Hauser,
  Wiruchnipawan and Burke}]{satterstrom2019thin}
\bibinfo{author}{Satterstrom, P.}, \bibinfo{author}{Polzer, J.T.},
  \bibinfo{author}{Kwan, L.B.}, \bibinfo{author}{Hauser, O.P.},
  \bibinfo{author}{Wiruchnipawan, W.}, \bibinfo{author}{Burke, M.},
  \bibinfo{year}{2019}.
\newblock \bibinfo{title}{Thin slices of workgroups}.
\newblock \bibinfo{journal}{Organizational Behavior and Human Decision
  Processes} \bibinfo{volume}{151}, \bibinfo{pages}{104--117}.
\bibitem[{Schuller et~al.(2012)Schuller, Valster, Eyben, Cowie and
  Pantic}]{schuller2012avec}
\bibinfo{author}{Schuller, B.}, \bibinfo{author}{Valster, M.},
  \bibinfo{author}{Eyben, F.}, \bibinfo{author}{Cowie, R.},
  \bibinfo{author}{Pantic, M.}, \bibinfo{year}{2012}.
\newblock \bibinfo{title}{Avec 2012: the continuous audio/visual emotion
  challenge}, in: \bibinfo{booktitle}{Proceedings of the 14th ACM international
  conference on Multimodal interaction}, \bibinfo{organization}{ACM}. pp.
  \bibinfo{pages}{449--456}.
\bibitem[{Sevier et~al.(2008)Sevier, Eldridge, Jones, Doss and
  Christensen}]{sevier2008observed}
\bibinfo{author}{Sevier, M.}, \bibinfo{author}{Eldridge, K.},
  \bibinfo{author}{Jones, J.}, \bibinfo{author}{Doss, B.D.},
  \bibinfo{author}{Christensen, A.}, \bibinfo{year}{2008}.
\newblock \bibinfo{title}{Observed communication and associations with
  satisfaction during traditional and integrative behavioral couple therapy}.
\newblock \bibinfo{journal}{Behavior therapy} \bibinfo{volume}{39},
  \bibinfo{pages}{137--150}.
\bibitem[{Smith(2017)}]{smith2017cyclical}
\bibinfo{author}{Smith, L.N.}, \bibinfo{year}{2017}.
\newblock \bibinfo{title}{Cyclical learning rates for training neural
  networks}, in: \bibinfo{booktitle}{2017 IEEE Winter Conference on
  Applications of Computer Vision (WACV)}, \bibinfo{organization}{IEEE}. pp.
  \bibinfo{pages}{464--472}.
\bibitem[{Srivastava et~al.(2014)Srivastava, Hinton, Krizhevsky, Sutskever and
  Salakhutdinov}]{srivastava2014dropout}
\bibinfo{author}{Srivastava, N.}, \bibinfo{author}{Hinton, G.},
  \bibinfo{author}{Krizhevsky, A.}, \bibinfo{author}{Sutskever, I.},
  \bibinfo{author}{Salakhutdinov, R.}, \bibinfo{year}{2014}.
\newblock \bibinfo{title}{Dropout: a simple way to prevent neural networks from
  overfitting}.
\newblock \bibinfo{journal}{The journal of machine learning research}
  \bibinfo{volume}{15}, \bibinfo{pages}{1929--1958}.
\bibitem[{Steiner and Rain(1989)}]{steiner1989immediate}
\bibinfo{author}{Steiner, D.D.}, \bibinfo{author}{Rain, J.S.},
  \bibinfo{year}{1989}.
\newblock \bibinfo{title}{Immediate and delayed primacy and recency effects in
  performance evaluation.}
\newblock \bibinfo{journal}{Journal of Applied Psychology}
  \bibinfo{volume}{74}, \bibinfo{pages}{136}.
\bibitem[{Stolcke(2002)}]{stolcke2002srilm}
\bibinfo{author}{Stolcke, A.}, \bibinfo{year}{2002}.
\newblock \bibinfo{title}{Srilm-an extensible language modeling toolkit}, in:
  \bibinfo{booktitle}{Seventh international conference on spoken language
  processing}.
\bibitem[{Thornton and Tamir(2017)}]{thornton2017mental}
\bibinfo{author}{Thornton, M.A.}, \bibinfo{author}{Tamir, D.I.},
  \bibinfo{year}{2017}.
\newblock \bibinfo{title}{Mental models accurately predict emotion
  transitions}.
\newblock \bibinfo{journal}{Proceedings of the National Academy of Sciences}
  \bibinfo{volume}{114}, \bibinfo{pages}{5982--5987}.
\bibitem[{Tseng et~al.(2017)Tseng, Baucom and Georgiou}]{tseng2017approaching}
\bibinfo{author}{Tseng, S.Y.}, \bibinfo{author}{Baucom, B.},
  \bibinfo{author}{Georgiou, P.}, \bibinfo{year}{2017}.
\newblock \bibinfo{title}{Approaching human performance in behavior estimation
  in couples therapy using deep sentence embeddings}, in:
  \bibinfo{booktitle}{Proceedings of Interspeech. August 2017}.
\bibitem[{Tseng et~al.(2016)Tseng, Chakravarthula, Baucom and
  Georgiou}]{tseng2016couples}
\bibinfo{author}{Tseng, S.Y.}, \bibinfo{author}{Chakravarthula, S.N.},
  \bibinfo{author}{Baucom, B.R.}, \bibinfo{author}{Georgiou, P.G.},
  \bibinfo{year}{2016}.
\newblock \bibinfo{title}{Couples behavior modeling and annotation using
  low-resource lstm language models.}, in: \bibinfo{booktitle}{INTERSPEECH},
  pp. \bibinfo{pages}{898--902}.
\bibitem[{Tseng et~al.(2018)Tseng, Li, Baucom and Georgiou}]{tseng2018honey}
\bibinfo{author}{Tseng, S.Y.}, \bibinfo{author}{Li, H.},
  \bibinfo{author}{Baucom, B.}, \bibinfo{author}{Georgiou, P.},
  \bibinfo{year}{2018}.
\newblock \bibinfo{title}{Honey, i learned to talk: Multimodal fusion for
  behavior analysis}, in: \bibinfo{booktitle}{Proceedings of the 2018 on
  International Conference on Multimodal Interaction},
  \bibinfo{organization}{ACM}. pp. \bibinfo{pages}{239--243}.
\bibitem[{Xia et~al.(2015)Xia, Gibson, Xiao, Baucom and
  Georgiou}]{xia2015dynamic}
\bibinfo{author}{Xia, W.}, \bibinfo{author}{Gibson, J.}, \bibinfo{author}{Xiao,
  B.}, \bibinfo{author}{Baucom, B.}, \bibinfo{author}{Georgiou, P.G.},
  \bibinfo{year}{2015}.
\newblock \bibinfo{title}{A dynamic model for behavioral analysis of couple
  interactions using acoustic features}, in: \bibinfo{booktitle}{Sixteenth
  Annual Conference of the International Speech Communication Association}.
\bibitem[{Xiao et~al.(2012)Xiao, Can, Georgiou, Atkins and
  Narayanan}]{xiao2012analyzing}
\bibinfo{author}{Xiao, B.}, \bibinfo{author}{Can, D.},
  \bibinfo{author}{Georgiou, P.G.}, \bibinfo{author}{Atkins, D.},
  \bibinfo{author}{Narayanan, S.S.}, \bibinfo{year}{2012}.
\newblock \bibinfo{title}{Analyzing the language of therapist empathy in
  motivational interview based psychotherapy}, in: \bibinfo{booktitle}{Signal
  \& Information Processing Association Annual Summit and Conference (APSIPA
  ASC), 2012 Asia-Pacific}, \bibinfo{organization}{IEEE}. pp.
  \bibinfo{pages}{1--4}.
\bibitem[{Zadeh et~al.(2018)Zadeh, Liang, Poria, Cambria and
  Morency}]{zadeh2018multimodal}
\bibinfo{author}{Zadeh, A.B.}, \bibinfo{author}{Liang, P.P.},
  \bibinfo{author}{Poria, S.}, \bibinfo{author}{Cambria, E.},
  \bibinfo{author}{Morency, L.P.}, \bibinfo{year}{2018}.
\newblock \bibinfo{title}{Multimodal language analysis in the wild: Cmu-mosei
  dataset and interpretable dynamic fusion graph}, in:
  \bibinfo{booktitle}{Proceedings of the 56th Annual Meeting of the Association
  for Computational Linguistics (Volume 1: Long Papers)}, pp.
  \bibinfo{pages}{2236--2246}.
\bibitem[{Zou(2007)}]{zou2007toward}
\bibinfo{author}{Zou, G.Y.}, \bibinfo{year}{2007}.
\newblock \bibinfo{title}{Toward using confidence intervals to compare
  correlations.}
\newblock \bibinfo{journal}{Psychological methods} \bibinfo{volume}{12},
  \bibinfo{pages}{399}.

\end{thebibliography}

\newpage

\appendix

\section{Algorithms}
\label{sec:append_alg}

\begin{algorithm}[!h]
\caption{Behavior Construct Similarity}
\begin{algorithmic}[1]
\For {each behavior $B_i$ in \textbf{B}}
	\For {each window length $L_k$ in \textbf{L}}
		\For {each functional $F_z$ in \textbf{F}}
			\State Initialize empty lists \textbf{G}, \textbf{H}
			\For {each session transcript $T_p$ in \textbf{T} containing $O_p$ words}
				\State Score $T_p$ at window length $L_k$ to get trajectory of scores \\ \hspace{3cm} \bm{$S^p_i$} = $\{S^1_i, S^2_i, \ldots S^{max(1, {O_p - L_k + 1})}_i\}$ for behavior $B_i$
				\State Compute aggregate score $F_z($\bm{$S^p_i$}$)$
				\State Append $F_z($\bm{$S^p_i$}$)$ to \textbf{G}
				\State Append ground-truth annotation $A_i^p$ to \textbf{H}
			\EndFor
			\State Compute Spearman Correlation $R_z(k)$ between \textbf{G} and \textbf{H}
		\EndFor
		\State $BCS_i(k) = \{ R_z(k) ~ \forall ~ z \}$
	\EndFor
\EndFor
\end{algorithmic}
\label{alg:bcs}
\end{algorithm}

\begin{algorithm}[!h]
\caption{Similarity-based Grouping of Behaviors}
\begin{algorithmic}[1]
\State Calculate $R_{global}$ = $R_\mathbb{G}$ as in Algorithm~\ref{alg:brc}
\For {number of clusters $U$ in \{$2, 3, 4$\}}
	\For {$10000$ random initializations}
		\State Run K-Means clustering on $R_{global}$ with $U$ clusters
		\State Store clustering scheme
	\EndFor
	\State Pick most frequently occurring unique clustering scheme ${cluster}_U$
	\State Calculate cluster size disparity ${dsp}_U$ = $Range$(cluster sizes in ${cls}_U$)
\EndFor
\State Pick Behavior Grouping ${cls}_{\argmin\limits_{U}{dsp}_U}$
\end{algorithmic}
\label{alg:beh_group}
\end{algorithm}

\begin{algorithm}[!h]
\caption{Behavior Relationship Consistency}
\begin{algorithmic}[1]
\For {each behavior $B_i$ in \textbf{B}}
	\For {each behavior $B_j$ in \textbf{B} $: j \neq i$}
		\For {each window length $L_k$ in \textbf{L}}
			\State Initialize empty lists \bm{$C_i$}, \bm{$C_j$}, \bm{$G_i$}, \bm{$G_j$}
			\For {each session transcript $T_p$ in \textbf{T} containing $O_p$ words}
				\State Score $T_p$ at window length $L_k$ to get trajectory of scores \\ \hspace{3cm} \bm{$S^p_i$} = $\{S^1_i, S^2_i, \ldots S^{max(1, {O_p - L_k + 1})}_i\}$ for $B_i$, \\ \hspace{3cm} \bm{$S^p_j$} = $\{S^1_j, S^2_j, \ldots S^{max(1, {O_p - L_k + 1})}_j\}$ for $B_j$
				\State Append \bm{$S^p_i$} to \bm{$C_i$}, \bm{$S^p_j$} to \bm{$C_j$}
				\State Append ground-truth annotation $A^p_i$ to \bm{$G_i$}, $A^p_j$ to \bm{$G_j$}
			\EndFor
			\State Compute Spearman Correlations $R_{\bm{C}}(i,j)$ between \bm{$C_i$} and \bm{$C_j$}, $R_{\bm{G}}(i,j)$ between \bm{$G_i$} and \bm{$G_j$}
			\State $BRC_{i,j}(k) = 1 - \frac{|R_{\bm{C}}(i,j) - R_{\bm{G}}(i,j)|}{2}$
		\EndFor
	\EndFor
\EndFor
\end{algorithmic}
\label{alg:brc}
\end{algorithm}

\section{Details of N-gram Model}
\label{sec:append_ngram_model}

\subsection{Model Description}
\label{ssec:append_ngram_desc}

Our modeling method, shown in Figure~\ref{fig:ngram_model}, assumes that behavior is rated on a scale from $1$ to $K$, where \textit{1} indicates the lowest degree (\dq{absence of behavior}) and $K$ indicates the highest degree (\dq{strong presence of behavior}).
We train $K-2$ pairs of LMs where the $r^{th}$ pair performs a binary classification of behavior belonging to Class \textit{0}, i.e. the range [\textit{1,r+1}], or Class \textit{1}, i.e. the range (\textit{r+1,K}].
Let's denote the behavior score of utterance $W$ as $x$ (which we want to estimate); then, the $r^{th}$ LM pair's N-gram likelihood probabilities can be expressed as:
\begin{align}
\label{eq:class_likelihoods}
P_0^r(W) & \equiv P(W \ | \ 1 \leq x \leq r\!+\!1)\\
P_1^r(W) & \equiv P(W \ | \ r\!+\!1 < x \leq K)
\end{align}

We compute binary posteriors from these binary likelihoods using Bayes Rule and assuming uniform priors as shown in Eqn.~\ref{eq:class_posteriors}.
Then, using Eqn.~\ref{eq:interval_posteriors}, the binary posteriors are converted into a probability mass function where the $r^{th}$ point represents the probability of behavior lying in the range [\textit{r,r+1}].
Finally, the behavior score $x$ for utterance $W$ is obtained by computing the expected value of this probability mass function, as in Eqn.~\ref{eq:expected_value}.

\begin{align}
P(x \leq r\!+\!1 \ |\ W) & \equiv P(1 \leq x \leq r\!+\!1 \ |\ W) \notag \\
& = \frac{P_0^r(W)P(1 \leq x \leq r\!+\!1)}{P_0^r(W)P(1 \leq x \leq r\!+\!1) + P_1^r(W)P(r\!+\!1 < x \leq K)} \label{eq:class_posteriors}\\
P(r < x \leq r\!+\!1 \ |\ W) & = P(x \leq r\!+\!1 \ |\ W) - P(x \leq r \ |\ W) \label{eq:interval_posteriors}
\end{align}
\begin{align}
x & = \sum_{r = 1}^{K-1}(r + \frac{1}{2})P(r < x \leq r\!+\!1 \ |\ W) \label{eq:expected_value}
\end{align}

\subsection{Training}
\label{ssec:append_ngram_train}

We implement Maximum Likelihood models through 3-gram LMs trained with Good-Turing discounting using the SRILM toolkit \citep{stolcke2002srilm}.
A leave-one-couple-out scoring scheme is used where models are trained on data from $Z - 1$ couples and subsequently used to score data from the $Z^{th}$ couple, in order to prevent overfiitting.

Ideally, these models would be trained at the same window length at which they would be tested, thereby resulting in five sets of models, one for each of the window lengths \{\textit{3, 10, 30, 50, 100}\} words.
However, it is not practically feasible to train N-gram models on sequences longer than 5 words, due to the \textit{curse of dimensionality} \citep{bengio2003neural}, where the amount of training data required increases exponentially with the order N.
Therefore, we instead train a single set of 3-gram models, i.e. at window length 3 words, and use them for testing at all the window lengths.
We show in the results in Sec.~\ref{ssec:res_winlen} that this train-test mismatch does not particularly bias our analysis towards always selecting 3 words as the appropriate window length.

\section{Details of Neural Model}
\label{sec:append_neural_model}

\subsection{Model Description}
\label{ssec:append_neural_desc}

Our model, shown in Figure~\ref{fig:neural_est}, is similar to the one used by Tseng et al. \citep{tseng2016couples} for classifying \textit{Negative} behavior from language, but with a few changes: the Long Short-Term Memory \citep{hochreiter1997long} unit is replaced with a Gated Recurrent Unit (GRU) \citep{cho2014properties} and the \textit{word2vec} \citep{mikolov2013distributed} embeddings are replaced by ELMo \citep{peters2018deep} embeddings.
Finally, while \citep{tseng2016couples} post-processed the system outputs using Support Vector Regression, we do not use such transformations since we are interested in analyzing the properties of the system outputs themselves.
Instead, we simply use a \textit{ReLU6} layer in order to ensure that our predictions are bounded, similar to the ground truth annotations \textbf{A}.

At runtime, given a windowed sequence of \textit{O} words $W=\{w_1,w_2,\ldots,w_O\}$ words from an observation window, we first mix, for each word, its ELMo embeddings using ELMo's weights.
This gives us a sequence of \textit{O} word embeddings.
This sequence is then passed to the GRU whose hidden state representations are of dimension \textit{V}.
Finally, the last hidden state of the GRU is passed to a fully connected layer, followed by a \textit{Relu6} \citep{krizhevsky2010convolutional} layer, resulting in a scalar value that represents the behavior score of the windowed sequence of words.

\subsection{ELMo Layer Weights}
\label{ssec:append_neural_elmo}

We use ELMo \citep{peters2018deep} embeddings which capture semantic and syntactic relations in a deep, contextual manner.
For every word, ELMo provides embeddings from 3 layers, each of dimension 1024, and a set of 3 mixing weights as well a scaling weight that can be trained in a task-specific manner.
As recommended by Peters et al. \citep{peters2018deep}, we obtain a single embedding for each word by taking the scaled, weighted sum of all 3 embeddings.
These mixing weights are softmax-normalized, similar to attention \citep{graves2013generating} weights, and, when trained, represent the relative importance of each layer in estimating behavior.

Generally, it has been observed in deep neural networks that lower layers tend to learn simpler representations such as edges in images and phrase-level information in text whereas higher layers tend to learn more complicated representations such as objects in images and semantic features in text \citep{olah2017feature,jawahar2019does}.
In particular, the higher layers in ELMo have been found to model complex characteristics of language such as polysemy and semantics better than the lower layers \citep{peters2018deep}.
Therefore, observing how these weights differ for each behavior group can illuminate how their linguistic characteristics are different.

\subsection{Training}
\label{ssec:append_neural_train}

The size of the GRU hidden representation \textit{V} is tuned to be either 10 or 100 and the sample minibatch size was set to 64.
Given a word sequence and the window length, zero padding is performed at the end wherever required.
Dropout \citep{srivastava2014dropout} of 0.2 is applied before the fully connected layer and all the model parameters are trained using backpropagation by optimizing \textit{L}1 loss in conjunction with Adam \citep{kingma2014adam} optimizer.
A separate model is trained for each behavior, without any shared parameters or layers, so as to ensure that the results are indicative of that behavior only.
We employ a 6-fold nested cross-validation setup where in every test fold, four folds are used to train the model while the fifth fold is used to optimize the model hyper-parameters, learning rate range, etc.
Similar to the setup with N-gram models, we ensure that no dyad appears in more than one fold.

Instead of performing grid search for tuning the learning rate, we use the Cyclical Learning Rate schedule as proposed by Smith \citep{smith2017cyclical}.
For each behavior and model configuration, we first perform a \dq{range test} to determine the minimum and maximum learning rates at which training remains stable.
We then cyclically vary the learning rate between its minimum and maximum value during training, saving a model checkpoint at the end of every epoch when the learning rate would be at its lowest.
Finally, at testing time, inspired by Huang et al. \citep{huang2017snapshot}, instead of using just the last or the best checkpoint, we use an ensemble average of all of them.

\begin{figure}[!h]
\centering
    \includegraphics[height=0.5\linewidth, width=1\linewidth]{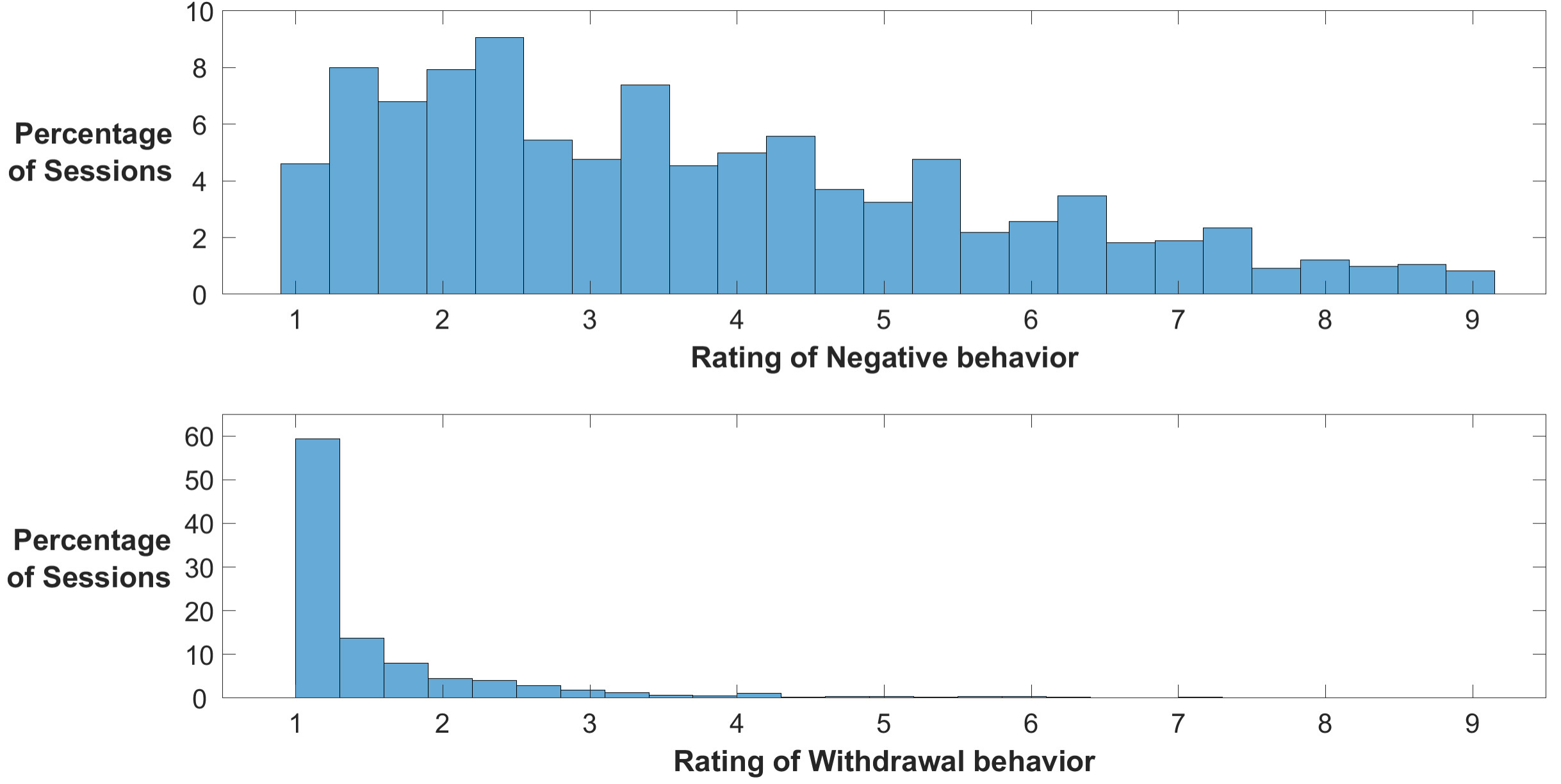}
    \caption{\textit{Distribution of expert ratings for different behaviors:}
    Behaviors such as \textit{Negative} (top) have an adequate number of data points (sessions) at all ratings, which results in minibatches with highly diverse ratings at training time.
    On the other hand, behavior such as \textit{Withdrawal} (bottom) are heavily skewed towards a single rating, with no data points at some ratings.
    This results in training minibatches where all the samples have very similar ratings, leading to a high chance of the \dq{dying ReLU} problem.
    }
    \label{fig:beh_dist}
\end{figure}

While we were able to analyze the Neural model at two window lengths: 3-word (i.e. 3 unrolled time steps) and 30-word (30 unrolled time steps), we were unable to do so at a long window, i.e. 100-word.
This was due to instability during training caused by the \dq{dying ReLU} problem in behaviors with highly skewed distributions of human ratings, such as \textit{Withdrawal}, shown in Figure~\ref{fig:beh_dist}.
Specifically, this problem occurred when randomly-shuffled minibatches ended up with nearly all its samples having the same rating as a result of the skewed distribution.
This would then result in a large gradient update that would cause the network weights to update in such a manner that the output ReLU6 layer would henceforth only output a 0 value, thereby rendering the Neural Model ineffective.
While we could resolve this problem at 3-word and 30-word window lengths by identifying stable learning rate ranges, we were unable to do the same at the 100-word window length.
As a result, we could not analyze the Neural Model at the 100-word window length.

\section{Intermediate Results for N-gram model}
\label{sec:append_res_ngram}

\subsection{Behavior Construct Similarity}
\label{ssec:append_res_ngram_bcs}

\begin{figure}[!h]
\centering
    \includegraphics[height=0.55\linewidth, width=1\linewidth]{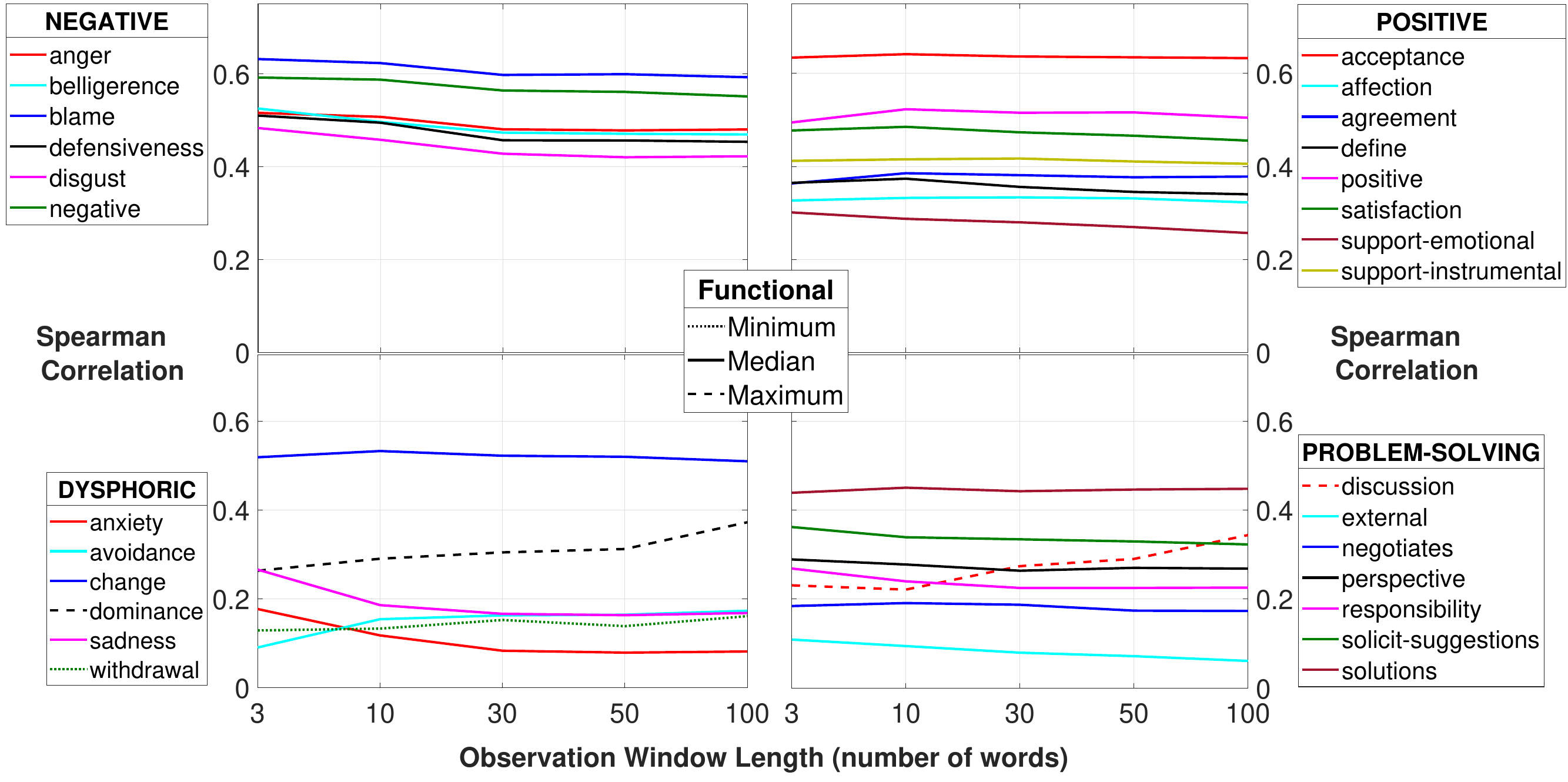}
    \caption{\textit{Behavior Construct Similarity for N-gram model:}
    Spearman Correlation between human annotations and functional-aggregated scores of N-gram model at different observation window lengths.
    All correlations are statistically significant $(p < 0.05)$}
    \label{fig:N-gram_bcs}
\end{figure}

Figure~\ref{fig:N-gram_bcs} shows the BCS of the N-gram model scores at the 5 observation window length values that were tested - \{\textit{3, 10, 30, 50, 100}\} words.
During the analysis procedure in Sec.~\ref{ssec:proc}, we set the BCS threshold $Y_1 = 0.59$ since the highest BCS, as can be seen in Figure~\ref{fig:N-gram_bcs}, is around 0.6.

Every behavior is represented by a trajectory and each point on the trajectory represents the Spearman Correlation between ground truth annotations and the aggregated model scores at that window length.
While we test three functionals for aggregation - \textit{minimum}, \textit{median} and \textit{maximum} - we only use the one that performed best, on average, across all window lengths for our analysis.
Hence, we only show the best performing functional for each behavior in the BCS plot; all correlations are statistically significant ($p < 0.05$).
We do not use the functional \textit{mean} because in some instances, we observe that the system's window-level scores were impulsive.
Fitting them to an $\alpha$-stable distributed results in an $\alpha \leq 1$, for which the \textit{mean} is undefined \citep{nolan2003stable}; the other three statistics, however, are still defined.

The best performing behaviors with the N-gram model are \textit{Acceptance} and \textit{Blame}, with BCS values greater than 0.6 at nearly all window lengths; hence, we use these as \textit{reliable behaviors} $B_{rel}$ during our analysis.
With respect to behavior groups, we see that the \textbf{Negative} behaviors are, on average, the best estimated ones, followed by \textbf{Positive} behaviors.
The BCS for \textbf{Problem-Solving} behaviors varies from moderate ($0.45$ for \textit{Solutions}) to extremely low ($0.06$ for \textit{External}).
Finally, with the exception of \textit{Change}, the BCS for \textbf{Dysphoric} behaviors is, in general, extremely low.
This matches previous studies which have found that behavioral constructs related to negative and positive affect tend to be estimated well from low-level lexical features.
From these results, we can now also see that they are, in fact, much better estimable than higher-level and more complex behaviors related to dysphoria and problem-solving.
This could be due to factors such as these behaviors not being expressed sufficiently in language or their expression in language, even if sufficient, being too complex to be modeled using N-gram phrases and simple statistics.

\begin{figure}[!h]
\centering
\includegraphics[height=0.55\linewidth, width=1\linewidth]{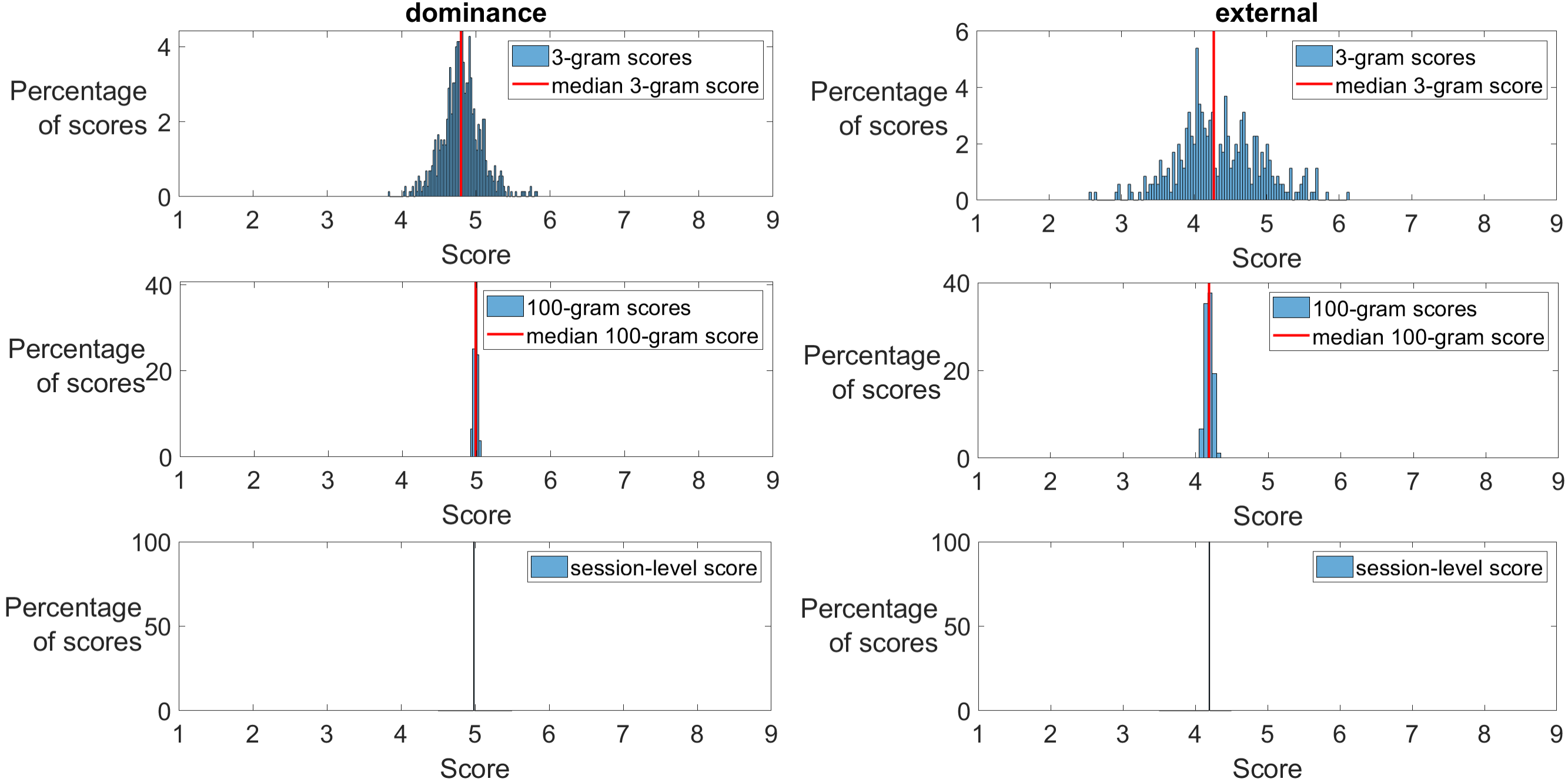}
\caption{\textit{Sample distribution of \emph{Dominance} and \emph{External} scores at window lengths 3, 100 and session-length:} In both behaviors, the median 3-gram as well as 100-gram scores are very similar to the session-level scores, possibly due to symmetrical distributions}
\label{fig:ses_scores}
\end{figure}

In evaluating the choice of functionals, \textit{median} appears to be the best aggregation method for nearly every behavior.
On the other hand, \textit{maximum} and \textit{minimum} perform best for some behaviors such as \textit{Dominance} and \textit{Withdrawal}.
In cases where \textit{median} is the best functional, we see that the BCS does not change much even from varying from the shortest window length to the longest one possible.
This, however, does not imply that all window lengths are equally appropriate for such behaviors.
As shown in Figure~\ref{fig:ses_scores}, the scores from the N-gram model tend to be symmetrically distributed, a pattern which was also reported in \citep{tseng2016couples}.
As a result, any change in scores resulting from changes in the window length would not be reflected by the median and, hence, the BCS would not change, giving the false impression that all windows are equally appropriate.
Hence, to further disambiguate this, we also check the Behavior Relationship Consistency (BRC) metric.

\subsection{Behavior Relationship Consistency}
\label{ssec:append_res_ngram_brc}

\begin{table}[h]
\centering
\footnotesize
\scalebox{0.925}{
\begin{tabular}{|c|c|ccccc|c|ccccc|}
\hline
\textbf{Reliable Behavior} & \multicolumn{6}{c|}{\textbf{Acceptance}} \\ \hline
\textbf{BRC Weight} &  & \begin{tabular}[c]{@{}c@{}}$\alpha(3)$\\ =0.501\end{tabular} & \begin{tabular}[c]{@{}c@{}}$\alpha(10)$\\ =0.507\end{tabular} & \begin{tabular}[c]{@{}c@{}}$\alpha(30)$\\ =0.516\end{tabular} & \begin{tabular}[c]{@{}c@{}}$\alpha(50)$\\ =0.514\end{tabular} & \begin{tabular}[c]{@{}c@{}}$\alpha(100)$\\ =0.516\end{tabular} \\ \hline
\textbf{Target Behavior} & \textbf{$Q^{*}$} & \textbf{$Q^{'}(3)$} & \textbf{$Q^{'}(10)$} & \textbf{$Q^{'}(30)$} & \textbf{$Q^{'}(50)$} & \textbf{$Q^{'}(100)$} \\ \hline
Discussion & 0.178 & -0.007 & 0.015 & 0.018 & 0.012 & 0.005 \\
External & 0.256 & 0.181 & 0.166 & 0.192 & 0.216 & 0.256 \\
Negotiates & 0.237 & 0.199 & 0.198 & 0.227 & 0.252 & 0.292 \\
Perspective & 0.099 & 0.068 & 0.041 & 0.021 & 0.006 & -0.022 \\
Responsibility & 0.328 & 0.198 & 0.221 & 0.257 & 0.284 & 0.329 \\
Solicit-suggestions & 0.379 & 0.297 & 0.305 & 0.349 & 0.381 & 0.434 \\
Solutions & 0.282 & 0.323 & 0.326 & 0.351 & 0.37 & 0.397 \\ \hline
Anger & -0.653 & -0.499 & -0.564 & -0.627 & -0.66 & -0.71 \\
Belligerence & -0.646 & -0.483 & -0.538 & -0.598 & -0.632 & -0.684 \\
Defensiveness & -0.564 & -0.446 & -0.487 & -0.548 & -0.585 & -0.644 \\
Disgust & -0.66 & -0.424 & -0.485 & -0.545 & -0.579 & -0.632 \\
Negative & -0.729 & -0.656 & -0.708 & -0.757 & -0.782 & -0.819 \\ \hline
Affection & 0.582 & 0.283 & 0.299 & 0.348 & 0.384 & 0.44 \\
Agreement & 0.347 & 0.269 & 0.259 & 0.284 & 0.307 & 0.34 \\
Define & 0.517 & 0.439 & 0.497 & 0.556 & 0.59 & 0.646 \\
Positive & 0.67 & 0.575 & 0.619 & 0.674 & 0.705 & 0.753 \\
Satisfaction & 0.563 & 0.456 & 0.486 & 0.538 & 0.572 & 0.624 \\
Support-emotional & 0.492 & 0.239 & 0.257 & 0.301 & 0.333 & 0.386 \\
Support-instrumental & 0.46 & 0.32 & 0.342 & 0.397 & 0.432 & 0.487 \\ \hline
Anxiety & -0.299 & -0.107 & -0.146 & -0.182 & -0.201 & -0.233 \\
Avoidance & -0.17 & 0.004 & -0.015 & -0.022 & -0.022 & -0.022 \\
Change & -0.474 & -0.47 & -0.528 & -0.579 & -0.606 & -0.653 \\
Dominance & -0.144 & -0.201 & -0.202 & -0.209 & -0.216 & -0.237 \\
Sadness & -0.131 & -0.059 & -0.069 & -0.075 & -0.074 & -0.07 \\
Withdrawal & -0.164 & 0.019 & - & 0.003 & 0.008 & 0.014 \\ \hline
\end{tabular}
}
\caption{\textit{Spearman Correlation between window-level scores of \textit{Acceptance} and target behaviors with the N-gram model:}
$Q^*$ and $Q^{'}$ refer to the correlations used to calculate the pair BRC in Eqn.~\ref{eqn:brc}.
$\alpha$ refers to the proportional weight used to calculate the individual BRC in Eqn.~\ref{eqn:brc_multi}.
All correlations are statistically significant ($p < 0.05$) unless marked as -}
\label{tab:brc_N-gram_acc}
\end{table}

\begin{table}[h]
\centering
\footnotesize
\scalebox{0.925}{
\begin{tabular}{|c|c|ccccc|c|ccccc|}
\hline
\textbf{Reliable Behavior} & \multicolumn{6}{c|}{\textbf{Blame}} \\ \hline
\textbf{BRC Weight} &  & \begin{tabular}[c]{@{}c@{}}$\alpha(3)$\\ =0.499\end{tabular} & \begin{tabular}[c]{@{}c@{}}$\alpha(10)$\\ =0.493\end{tabular} & \begin{tabular}[c]{@{}c@{}}$\alpha(30)$\\ =0.484\end{tabular} & \begin{tabular}[c]{@{}c@{}}$\alpha(50)$\\ =0.486\end{tabular} & \begin{tabular}[c]{@{}c@{}}$\alpha(100)$\\ =0.484\end{tabular} \\ \hline
\textbf{Target Behavior} & \textbf{$Q^{*}$} & \textbf{$Q^{'}(3)$} & \textbf{$Q^{'}(10)$} & \textbf{$Q^{'}(30)$} & \textbf{$Q^{'}(50)$} & \textbf{$Q^{'}(100)$} \\ \hline
Discussion & 0.085 & -0.21 & -0.191 & -0.174 & -0.161 & -0.146 \\
External & -0.073 & 0.212 & 0.176 & 0.12 & 0.08 & 0.015 \\
Negotiates & -0.083 & 0.181 & 0.138 & 0.076 & 0.033 & -0.036 \\
Perspective & - & 0.141 & 0.124 & 0.123 & 0.127 & 0.139 \\
Responsibility & -0.239 & 0.176 & 0.102 & 0.029 & -0.016 & -0.092 \\
Solicit-suggestions & -0.23 & 0.083 & 0.013 & -0.067 & -0.116 & -0.195 \\
Solutions & -0.184 & 0.006 & -0.057 & -0.117 & -0.153 & -0.207  \\ \hline
Anger & 0.673 & 0.736 & 0.739 & 0.762 & 0.778 & 0.806 \\
Belligerence & 0.677 & 0.738 & 0.735 & 0.756 & 0.773 & 0.8 \\
Defensiveness & 0.522 & 0.639 & 0.635 & 0.666 & 0.689 & 0.729 \\
Disgust & 0.69 & 0.707 & 0.708 & 0.729 & 0.745 & 0.773 \\
Negative & 0.693 & 0.774 & 0.784 & 0.809 & 0.825 & 0.852  \\ \hline
Affection & -0.352 & 0.152 & 0.081 & -0.004 & -0.058 & -0.143 \\
Agreement & -0.295 & 0.118 & 0.072 & 0.008 & -0.032 & -0.097 \\
Define & -0.353 & -0.536 & -0.552 & -0.594 & -0.622 & -0.674 \\
Positive & -0.547 & -0.201 & -0.297 & -0.39 & -0.44 & -0.522 \\
Satisfaction & -0.537 & -0.095 & -0.182 & -0.27 & -0.32 & -0.401 \\
Support-emotional & -0.326 & 0.181 & 0.119 & 0.042 & -0.008 & -0.086 \\
Support-instrumental & -0.343 & 0.089 & 0.004 & -0.089 & -0.143 & -0.23  \\ \hline
Anxiety & 0.17 & 0.425 & 0.408 & 0.413 & 0.417 & 0.426 \\
Avoidance & 0.085 & 0.333 & 0.307 & 0.287 & 0.273 & 0.253 \\
Change & 0.7 & 0.703 & 0.704 & 0.726 & 0.743 & 0.77 \\
Dominance & 0.293 & 0.174 & 0.224 & 0.234 & 0.24 & 0.254 \\
Sadness & 0.198 & 0.394 & 0.354 & 0.328 & 0.312 & 0.288 \\
Withdrawal & - & 0.293 & 0.268 & 0.241 & 0.224 & 0.2 \\ \hline
\end{tabular}
}
\caption{\textit{Spearman Correlation between window-level scores of \textit{Blame} and target behaviors with the N-gram model:}
$Q^*$ and $Q^{'}$ refer to the correlations used to calculate the pair BRC in Eqn.~\ref{eqn:brc}.
$\alpha$ refers to the proportional weight used to calculate the individual BRC in Eqn.~\ref{eqn:brc_multi}.
All correlations are statistically significant ($p < 0.05$) unless marked as -}
\label{tab:brc_N-gram_bla}
\end{table}

Tables~\ref{tab:brc_N-gram_acc} and \ref{tab:brc_N-gram_bla} display the Spearman Correlations between different target behaviors and \textit{Acceptance} and \textit{Blame} respectively, which are the \textit{reliable behaviors} for the Ngram model.
$Q^*$ represents the \dq{true correlation} i.e. the correlation between the ground-truth annotations.
$Q^{'}(L)$ represents the correlation between the model's scores at window length $L$.
Negative values signify that the behaviors are dissimilar or opposite whereas positive values signify that the two behaviors are similar.
For example, we can see from Table~\ref{tab:beh_list} that \textit{Anger} is similar to \textit{Blame} but dissimilar to \textit{Acceptance}.
This is reflected in the $Q^*$ for \textit{Acceptance} and \textit{Anger} in Table~\ref{tab:brc_N-gram_acc} which is -0.653 since they are dissimilar whereas the $Q^*$ for \textit{Blame} and \textit{Anger} in Table~\ref{tab:brc_N-gram_bla} is 0.673 since they are similar.

Next, we calculate the normalized weights $\alpha$ at each window length for the two reliable behaviors \textit{Acceptance} and \textit{Blame} as shown in Eqn.\ref{eqn:rel_weights}; these are shown in the second row of Tables~\ref{tab:brc_N-gram_acc} and \ref{tab:brc_N-gram_bla} respectively.
Finally, we plug in $Q^*$, $Q^{'}$ and $\alpha$ into Eqn.~\ref{eqn:brc_multi} to obtain the BRC of each target behavior at every window length.
During the analysis procedure in Sec.~\ref{ssec:proc}, we set the BRC threshold $Y_2 = 0.95$ in order to be as close to 1 as practically possible.
This now enables us to observe changes in the quality of the N-gram model scores which are otherwise not reflected in the BCS.
For instance, for the behavior \textit{Solutions}, the BCS is nearly the same, around 0.45 at all the window lengths.
However, when compared to its ground truth correlations with \textit{Acceptance} and \textit{Blame} (0.282 and -0.184 respectively), we see that the N-gram score correlations at 50 words (0.37 and -0.153 respectively) are more similar to them than those at 3 words (0.323 and 0.006 respectively).
Therefore, based on this, we can conclude that a window length of 50 words is more appropriate for scoring the behavior \textit{Solutions} than 3 words.

\section{Intermediate Results for Neural model}
\label{sec:append_res_neural}

\subsection{Behavior Construct Similarity}
\label{ssec:append_res_neural_bcs}

\begin{figure}[!h]
\centering
    \includegraphics[height=0.525\linewidth, width=1\linewidth]{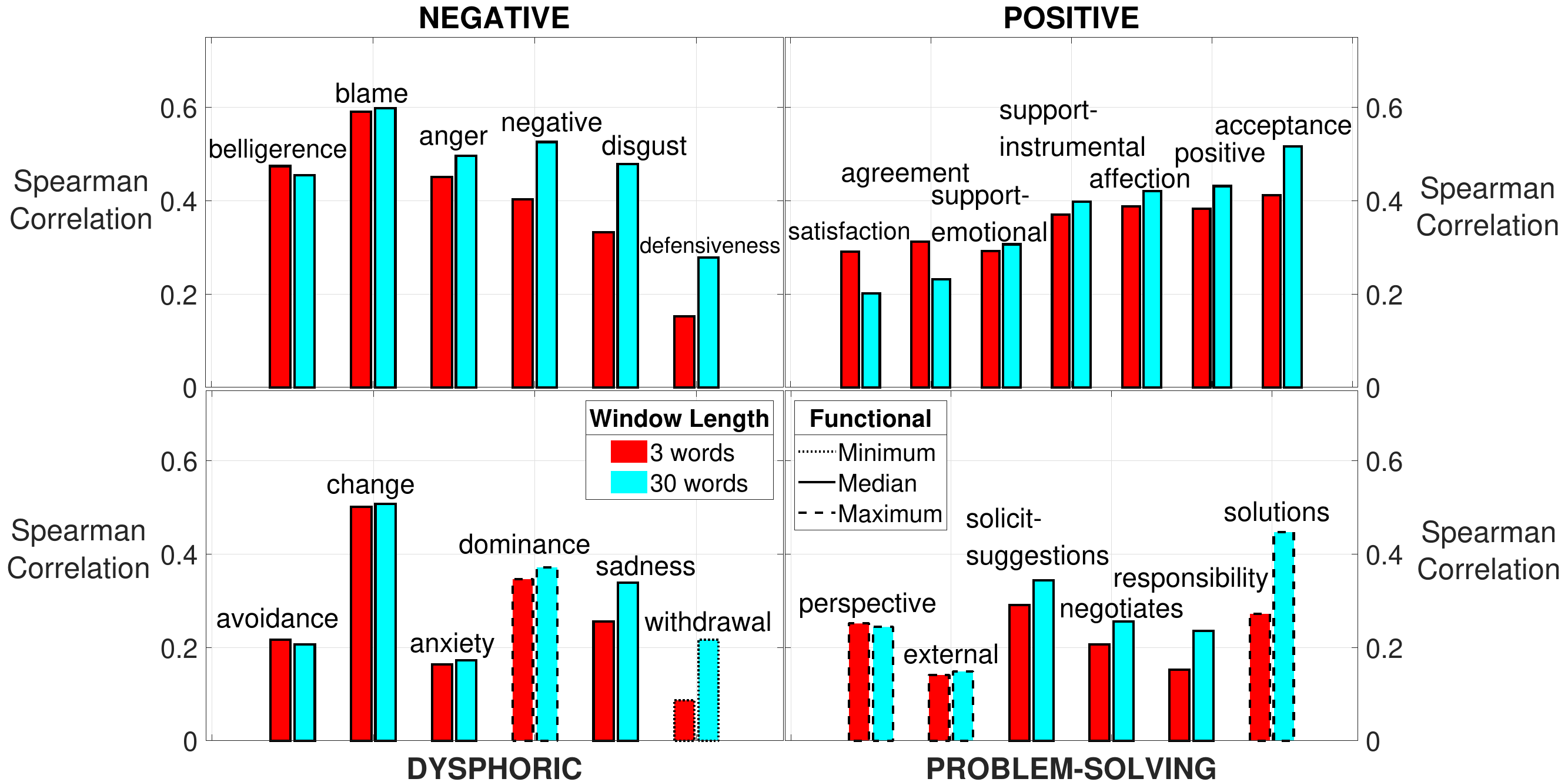}
    \caption{\textit{Behavior Construct Similarity for Neural model:}
    Spearman Correlation between human annotations and functional-aggregated scores of Neural model at different observation window lengths.
    All correlations are statistically significant $(p < 0.05)$}
    \label{fig:neural_bcs}
\end{figure}

Figure~\ref{fig:neural_bcs} shows the BCS of the Neural model scores at the two observation window lengths tested, $3$ and $30$ words.
For each behavior, a bar represents the Spearman Correlation between ground truth annotations and the aggregated Neural model score at that window length.
Similar to the N-gram model, we used the best performing functional, on average, for our analysis and display it for each behavior in the BCS plot; all correlations are statistically significant ($p < 0.05$).
Similar to the N-gram model, for the analysis procedure in Sec.~\ref{ssec:proc}, we set the BCS threshold $Y_1 = 0.59$ since the highest BCS, as can be seen in Figure~\ref{fig:neural_bcs}, is around 0.6.

The best performing behavior with the Neural model is \textit{Blame}; hence, it is used as the \textit{reliable behavior} in our analysis.
Once again, we see that the best estimated behaviors belong to the \textbf{Negative} group, followed by \textbf{Positive}.
Interestingly, however, we observe low-to-moderate BCS for both \textbf{Dysphoric} behaviors, which are generally subtle and non-verbal, as well as \textbf{Problem-Solving} behaviors, which are generally verbose.
This shows that the contextual embeddings of ELMo in conjunction with the long-term, non-linear processing of the GRU are able to handle both scenarios' diverse linguistic requirements equally well.

We also see that the BCS for some behaviors changes noticeably as the window length increases from 3 to 30 words.
Technically, this variation can be attributed to not just the change in window length but also the quality of the model trained at that window length.
However, since we tuned for the best Neural model at each window length, we assume that the quality of training is similar across window lengths and that the variation in BCS is mostly due to the change in window length.

\subsection{Behavior Relationship Consistency}
\label{ssec:append_res_neural_brc}

\begin{table}[!h]
\centering
\scalebox{0.8}{
\begin{tabular}{|c|c|cc|}
\hline
\textbf{Reliable Behavior} & \multicolumn{3}{c|}{\textbf{Blame}} \\ \hline
\textbf{Target Behavior} & \textbf{$Q^{*}$} & \textbf{$Q^{'}(3)$} & \textbf{$Q^{'}(30)$} \\ \hline
External & -0.073 & -0.18 & -0.134 \\
Negotiates & -0.083 & -0.381 & -0.263 \\
Perspective & - & 0.063 & 0.0902 \\
Responsibility & -0.239 & -0.311 & -0.346 \\
Solicit-suggestions & -0.23 & -0.388 & -0.42 \\
Solutions & -0.184 & -0.306 & -0.356 \\ \hline
Anger & 0.673 & 0.678 & 0.659 \\
Belligerence & 0.677 & 0.605 & 0.55 \\
Defensiveness & 0.522 & 0.408 & 0.454 \\
Disgust & 0.69 & 0.634 & 0.421 \\
Negative & 0.693 & 0.732 & 0.457 \\ \hline
Acceptance & -0.75 & -0.656 & -0.618 \\
Affection & -0.352 & -0.481 & -0.528 \\
Agreement & -0.295 & -0.464 & -0.347 \\
Positive & -0.547 & -0.61 & -0.59 \\
Satisfaction & -0.537 & -0.465 & -0.464 \\
Support-emotional & -0.326 & -0.541 & -0.358 \\
Support-instrumental & -0.343 & -0.571 & -0.39 \\ \hline
Anxiety & 0.171 & 0.369 & 0.349 \\
Avoidance & 0.085 & - & -0.005 \\
Change & 0.7 & 0.76 & 0.725 \\
Dominance & 0.293 & 0.394 & 0.425 \\
Sadness & 0.198 & 0.204 & 0.177 \\
Withdrawal & - & -0.118 & -0.035 \\ \hline
\end{tabular}
}
\caption{\textit{Spearman Correlation between window-level scores of \textit{Blame} and target behaviors with the Neural model:}
$Q^*$ and $Q^{'}$ refer to the correlations used to calculate the pair BRC in Eqn.~\ref{eqn:brc}.
Since we have only one , $\alpha = 1$ in Eqn.~\ref{eqn:brc_multi}.
All correlations are statistically significant ($p < 0.05$) unless marked as -}
\label{tab:brc_neural}
\end{table}

Table~\ref{tab:brc_neural} shows the Spearman Correlations between target behaviors and \textit{Blame}, reliable behavior in the Neural model.
$Q^*$ represents the \dq{true correlation} i.e. the correlation between the ground-truth annotations.
$Q^{'}(L)$ represents the correlation between the model's scores at window length $L$.
Since we have only one \textit{reliable behavior}, we calculate the BRC using Eqn.~\ref{eqn:brc_multi} with $\alpha = 1$.
Similar to the N-gram model, for the analysis procedure in Sec.~\ref{ssec:proc}, we set the BRC threshold $Y_2 = 0.95$.

Then, using both BCS and BRC, we analyze the Neural model scores of all the behaviors, the results of which are shown in Figure~\ref{fig:final_analysis}.
Since we are comparing just two window lengths, 3 and 30 words, the resolution of our analysis here is slightly coarse and doesn't necessarily reflect exhaustive trends.
For instance, a behavior that actually requires 10-word windows might perform better at 30 words than at 3 words simply because of the increased context and not because it is best observed at 30 words.
Hence, the window length results in the Neural model should be interpreted in a relative light, i.e. short window vs \textit{longer} window.

\subsection{ELMo Layer Weights}
\label{ssec:append_res_neural_elmo}

The 3 layers in ELMo, from bottom to top, are the input token layer and 2 bidirectional language model (biLM) layers.
Figure~\ref{fig:neural_weights} displays the mixing weights of each layer, averaged over all model checkpoints, from each test fold and for each behavior in a group.

We see that \textbf{Problem-Solving} behaviors tend to predominantly use information mostly from the top layer, followed by the middle layer.
This means that the models used to estimate these behaviors rely mostly on the biLM representations which, as we noted earlier, pertain to complex and high-level characteristics of language.

While \textbf{Negative} and \textbf{Positive} behaviors similarly place heavy emphasis on the top layer, they assign a much larger weight to the bottom layer, which typically encodes word-level features; this matches the notion that it is possible to express them using short expressions.
Finally, we see that \textbf{Dysphoric} behaviors assign similar weights to all 3 layers, implying that we need to extract information from all 3 aspects of language in order to capture them.

\end{document}